\documentclass[journal]{IEEEtran}

\usepackage[utf8]{inputenc} % allow utf-8 input
\usepackage[T1]{fontenc}    % use 8-bit T1 fonts
\usepackage{hyperref}       % hyperlinks
\usepackage{url}            % simple URL typesetting
\usepackage{booktabs}       % professional-quality tables
\usepackage{amsfonts}       % blackboard math symbols
\usepackage{nicefrac}       % compact symbols for 1/2, etc.
\usepackage{microtype}      % microtypography

\usepackage{graphicx}
\usepackage{amsmath,amssymb} % define this before the line numbering.}
\usepackage{color,xcolor}
\usepackage[ruled,vlined]{algorithm2e}
\usepackage{multirow}

\newcommand{\eg}{\textit{e.g.}}
\newcommand{\ie}{\textit{i.e.}}
\newcommand{\etal}{\textit{et al.}}
\newcommand{\argmax}{\mathop{\arg\max}}
\newcommand{\Sdomain}{\mathcal{S}}
\newcommand{\Tdomain}{\mathcal{T}}
\newcommand{\cdl}{\mathcal{L}_{cdl}}
\newcommand{\yyp}{\mathop{I}{(y_{i}, y_{\hat{i}})}}
\newcommand{\yym}{\beta_{\hat{i}}}
\providecommand{\keywords}[1]
{
  \small	
  \textbf{\textit{Index terms---}} #1
}

\title{Effective Label Propagation for Discriminative Semi-Supervised Domain Adaptation}

\author{Zhiyong~Huang, Kekai~Sheng, Weiming~Dong,~\IEEEmembership{Member,~IEEE,} Xing~Mei,~\IEEEmembership{Member,~IEEE,} Chongyang~Ma,~\IEEEmembership{Member,~IEEE,} Feiyue~Huang, Dengwen~Zhou, Changsheng~Xu,~\IEEEmembership{Fellow,~IEEE} 
\IEEEcompsocitemizethanks{
\IEEEcompsocthanksitem Z. Huang and D. Zhou are with School of Control and Computer Engineering, North China Electric Power University. E-mail: \{1182227193, zdw\}@ncepu.edu.cn.
\IEEEcompsocthanksitem W. Dong and C. Xu are with NLPR, Institute of Automation, Chinese Academy of Sciences and School of Artificial Intelligence, University of Chinese Academy of Sciences, Beijing, China. E-mail: \{weiming.dong, changsheng.xu\}@ia.ac.cn.
\IEEEcompsocthanksitem K. Sheng and F. Huang are with Youtu lab, Tencent, Shanghai, China. E-mail: \{saulsheng, garyhuang\}@tencent.com.
\IEEEcompsocthanksitem X. Mei is with Bytedance Inc, Beijing, China. E-mail:xing.mei@bytedance.com.
\IEEEcompsocthanksitem C. Ma is with Kuaishou Technology, Beijing, China. E-mail: chongyangma@kwai.com.
}
}

\begin{document}

% The paper headers
\markboth{IEEE TRANSACTIONS ON IMAGE PROCESSING, VOL. XXX, 2021}%
{Huang \MakeLowercase{\textit{et al.}}: Effective Label Propagation for Discriminative Semi-Supervised Domain Adaptation}

\maketitle

\begin{abstract}
Semi-supervised domain adaptation (SSDA) methods have demonstrated great potential in large-scale image classification tasks when massive labeled data are available in the source domain but very few labeled samples are provided in the target domain.
Existing solutions usually focus on feature alignment between the two domains while paying little attention to the discrimination capability of learned representations in the target domain.
In this paper, we present a novel and effective method, namely Effective Label Propagation (ELP), to tackle this problem by using effective inter-domain and intra-domain semantic information propagation.
For inter-domain propagation, we propose a new cycle discrepancy loss to encourage consistency of semantic information between the two domains.
For intra-domain propagation, we propose an effective self-training strategy to mitigate the noises in pseudo-labeled target domain data and improve the feature discriminability in the target domain.
As a general method, our ELP can be easily applied to various domain adaptation approaches and can facilitate their feature discrimination in the target domain.
% To verify its versatility, we evaluate the proposed ELP in two typical adaptation cases: semi-supervised and close-set unsupervised domain adaptation.
Experiments on Office-Home and DomainNet benchmarks show that ELP consistently improves the classification accuracy of mainstream SSDA methods by $2\% \sim 3 \%$.
Additionally, ELP also improves the performance of UDA methods as well ($81.5 \%$ vs $86.1 \%$), based on UDA experiments on the VisDA-2017 benchmark.
Our source code and pre-trained models will be released soon.

\end{abstract}

\section{Introduction}

%%%%%
% \paragraph{The significance}
Deep convolutional neural networks (CNNs) have significantly advanced state-of-the-art in large-scale image classification on public datasets such as ImageNet~\cite{deng2009imagenet} and Open Image~\cite{kuznetsova2018openimage}. Massive labeled training data are essential for the superior performance of CNNs, but they are not always available in a new domain practically. One popular approach to tackle this problem is \textit{domain adaptation} (DA)~\cite{long2015mmd,csurka2017domain,long2018conditional,saito2019semi,liang2020shot}.
The goal of DA is to leverage labeled data from a source domain to boost unsupervised learning (UDA) or semi-supervised learning (SSDA) in a new but related target domain.
Over the past decades, DA has been successfully applied in many scenarios, such as image classification~\cite{csurka2017domain,long2018conditional,li2018domain,pan2019transferrableproto}, object detection~\cite{xu2020explore}, semantic segmentation~\cite{zhou2020affinity}, and person re-identification~\cite{deng2018image}.

\begin{figure}
    \centering
    \includegraphics[width=0.96\linewidth]{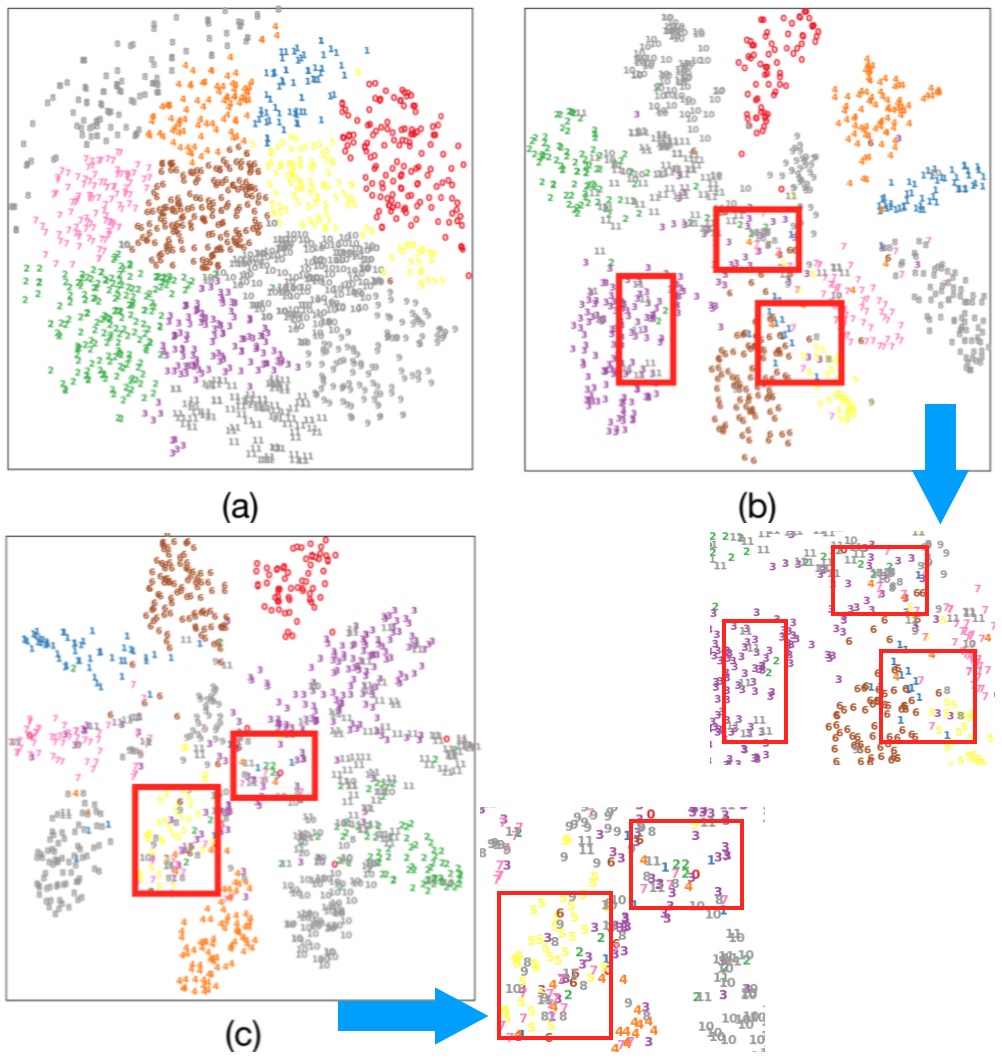}
    \caption{A t-SNE visualization~\cite{maaten2008tsne} of the embedded features for the \emph{synthetic-to-real} task from the VisDA-2017 dataset~\cite{peng2017visda}. Features from classes $0$ to $11$ are marked with different colors. (a) Feature distribution in the source domain. Feature distribution in the target domain with (b) ``Drop to Adapt'' (DTA)~\cite{lee2019drop} (a recent UDA method) , and (c) Minimax Entropy (MME)~\cite{saito2019semi} (a recent SSDA method). %, and (d) the proposed ELP method.
    This figure is best viewed in the electronic version.}
    \label{fig:PoorDiscriminationOfSOTA}
\end{figure}

\begin{figure*}
    \centering
    \includegraphics[width=0.86\linewidth]{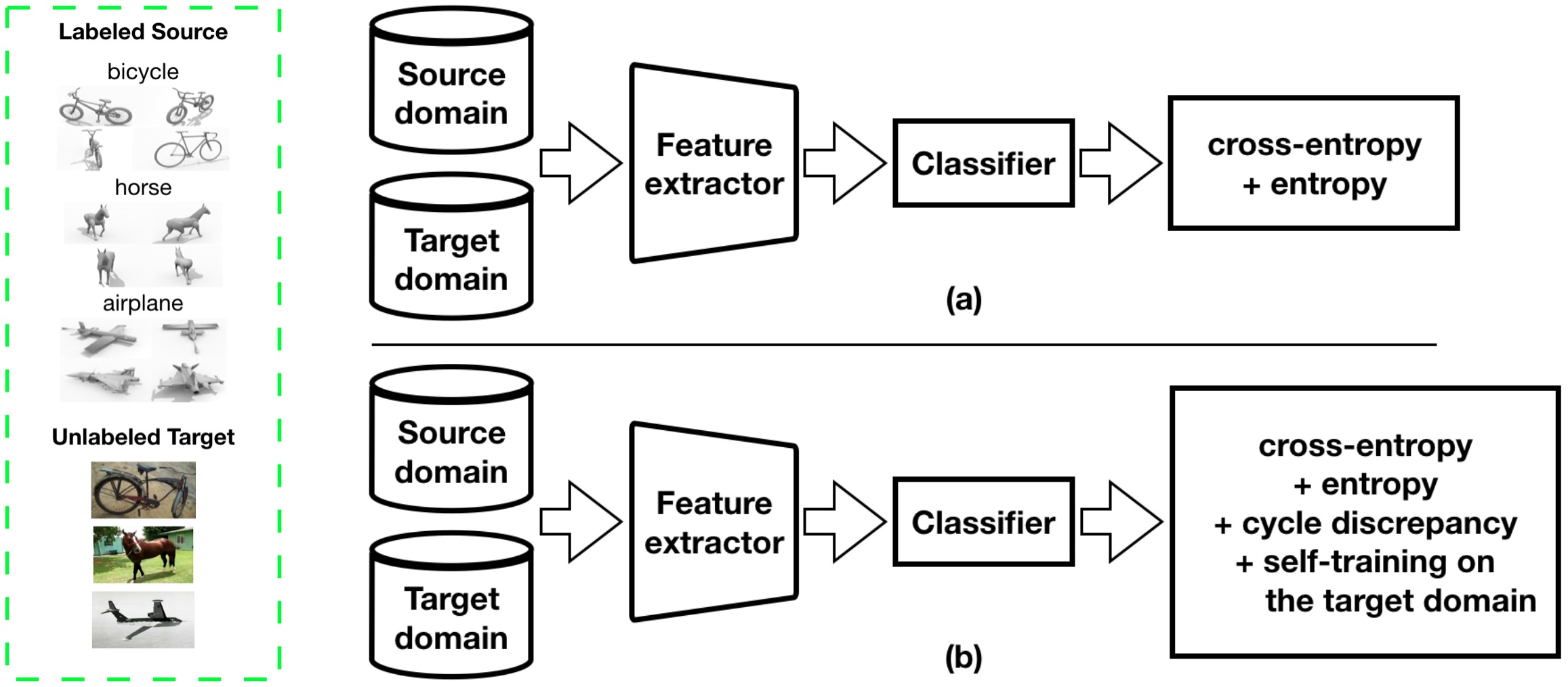}
    \caption{The schematic illustration of the difference between (a) the existing method MME~\cite{saito2019semi} and (b) the proposed ELP framework.}
    \label{fig:PreviousVSOurs}
\end{figure*}

%%%%%
% \paragraph{Related works and big yet-unsolved solvable challenges}
A central challenge of domain adaptation is the gap of the feature distributions between the two domains. Conventional UDA and SSDA methods focus on aligning these distributions with adversarial training~\cite{long2015mmd,csurka2017domain,saito2018MCD}.
More recent SSDA methods jointly minimize the task loss and the domain gap with subspace learning~\cite{yao2015SDASL}, consistency regularization~\cite{kumar2010EA++,french2018SE,xu2019dSNE}, or entropy-based loss design~\cite{saito2019semi}. These methods intend to learn transferable visual features for better classification accuracies on various DA datasets.
Nevertheless, it is arguable that the feature transferability is not equal to the feature discriminability, and the classification accuracies on the target domain still remain unsatisfactory~\cite{chen2019transferability,liang2020shot}.
We illustrate the problem with an example (Synthetic to Real) from the VisDA-2017 dataset~\cite{peng2017visda}, as illustrated in Fig.~\ref{fig:PoorDiscriminationOfSOTA}:
the representations extracted by existing UDA (Fig.~\ref{fig:PoorDiscriminationOfSOTA}(b)) and SSDA methods (Fig.~\ref{fig:PoorDiscriminationOfSOTA}(c)) can fail to form clear decision boundaries in some target domain regions (marked with red rectangles).
%%% 如果觉得句子写得有逻辑、有争议，请明显地标记出来。像下面这样就改一个词，很难发现。。。
%%% 多谢老师
% It is controversial that the unsatisfactory discriminability has its origin from the fact that the task-specific label information might not have been fully exploited by existing DA methods.
That is to say, one ideal DA method should take into consideration the transferability and the discriminability of learned representations at the same time.

%%%%%
% \paragraph{Motivation}
There have been a few attempts in improving the discrimination of learned representations on the target domains.
Representative methods are ~\cite{xu2019dSNE,zou2019confidence,liang2020shot}.
Despite their positive effects, the classification accuracies on large-scale datasets (e.g., for VisDA-2017~\cite{peng2017visda}, please refer to Table~\ref{tab:ResultsOnVisDA2017} in Section~\ref{sec:Experiments}) are still unsatisfactory, which might suffer from the noise of pseudo-labeled data on the target domain.
In this paper, we introduce a new learning scheme with effective label propagation to propagate semantic-aware information from the source domain to the target domain, reinforce the feature discriminability on the target domain with an effective self-training strategy.
In this way, we eventually promote the classification performance on the target domain.

%%%%%
% \paragraph{Proposed Methodology}
Specifically, we propose a novel framework, namely Effective Label Propagation (ELP), to propagate label information both across the two domains and within the target domain.
Fig.~\ref{fig:PreviousVSOurs}(b) shows the framework of our ELP.
We first propose a cycle discrepancy loss (CDL) which requires the feature extractor to explain labeled samples in the source domain with unlabelled samples from the target domain and encourage label consistency between the two domains.
With CDL, we encourage the label consistency in the representation space between the two domains.
Besides, we conduct intra-domain propagation ($\ie$ an effective self-training scheme) to enhance the discrimination of features on the target domain.
To combat against the noises in pseudo-labeled instances in the target domain, we propose some simple yet effective strategies: memory bank, dynamic threshold strategy, and balanced pseudo-label aggregation.
% Overall, ELP works in a two-stage process.
We evaluate ELP on two domain adaptation benchmarks, including Office-Home~\cite{venkateswara2017deep} (small-scale), DomainNet~\cite{peng2019moment} (large-scale), and the results in Section~\ref{sec:Experiments} demonstrate the effectiveness of the proposed ELP.
Besides, we also find that our ELP helps promote the classification accuracies of \textit{state-of-the-art} UDA methods ($\eg$ DTA~\cite{lee2019drop}) building on the UDA experiments on VisDA-2017 benchmark.
These experimental results further verify the versatility of ELP in domain adaptation.
% The experimental results in Section~\ref{sec:Experiments} demonstrates that our method consistently improves the classification performance of \textit{state-of-the-art} domain adaptation approaches in two typical adaptation scenarios, both SSDA and UDA.

%%%%%
%\paragraph{Summary}
In summary, the technical contributions of our ELP are:
\begin{itemize}
    \item To propagate semantic-aware information from the source domain and the target domain effectively, we propose a novel loss function based on cycle discrepancy.
    
    \item To improve feature discriminability in the target domain further, we propose an effective self-training strategy: memory bank, dynamic threshold strategy, and balanced pseudo-label aggregation.
    
    \item Extensive experiments and ablation studies on several domain adaptation benchmarks and two typical adaptation scenarios (SSDA + UDA) demonstrate the effectiveness and versatility of the proposed ELP.
\end{itemize}

\section{Related work}

%%%%% for the sake of layout
\begin{figure*}
    \centering
    \includegraphics[width=0.81\linewidth]{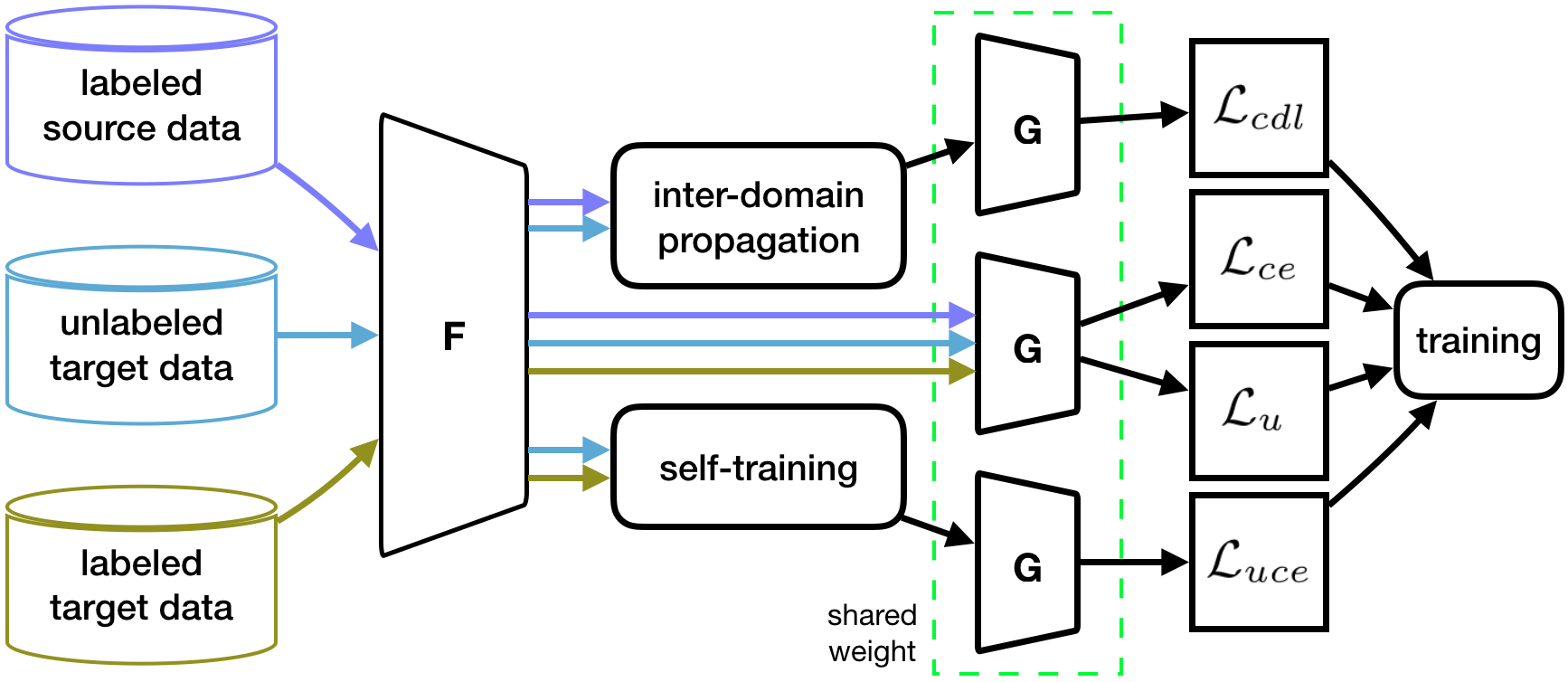}
    \caption{The illustration of the proposed ELP. ELP extracts features with a CNN architecture $F$, directs these features from different domains to different modules, calculates different losses $\mathcal{L}_{cdl}, \mathcal{L}_{ce}, \mathcal{L}_{u}, \mathcal{L}_{uce}$ with a classifier $G$ and minimizes the overall loss in the training process. The key contributions of ELP lie in two aspects.
    First, we introduce a new cycle discrepancy loss $\mathcal{L}_{cdl}$ that conducts inter-domain propagation among the two domains.
    Second, we reinforce the feature discriminability on the target domain with self-training. %and the cross entropy loss $\mathcal{L}_{uce}$. 
    %This figure is best viewed in the electronic version.
    }
    \label{fig:Framework}
\end{figure*}

%%%%%
\subsection{Domain Adaptation.}
Over the past decades, a number of domain adaptation (DA) methods have been proposed to minimize domain discrepancy and obtain domain-invariant features~\cite{french2018SE,long2015mmd,long2018conditional,pan2019transferrableproto,xu2019AFN,liang2020shot,xu2019adversarial}. A comprehensive survey of all DA methods is beyond the scope of this paper, and we mainly focus on recent semi-supervised domain adaptation (SSDA) methods that use a few labeled samples in the target domain. Yao \etal~\cite{yao2015SDASL} proposed a subspace learning framework (SDASL) which projects samples from both domains into this subspace and learns classifiers with several regularization strategies. Xu \etal~\cite{xu2019dSNE} used stochastic neighborhood embedding (d-SNE) to transform features into a common latent space for few-shot supervised learning and improve the feature discrimination on the target domain via metric learning. Recently, Saito \etal~\cite{saito2019semi} proposed a minimax entropy approach (MME) which iterative updates the task classifier and the feature extractor in a min-max entropy training scheme.

Different from the aforementioned counterparts, the proposed method explicitly leverages semantic-aware information via domain propagation for enhanced feature discriminability on the target domain, which has not been fully exploited in the research field of SSDA. For a better comparison of different approaches, we list their core characteristic information in Table~\ref{tab:MainstreamOfDAMethods}. Numerical results in Section~\ref{sec:Experiments} will verify the effectiveness of the proposed novel learning scheme.

\begin{table}
    \centering
    \caption{Comparison of state-of-the-art and related methodologies with the proposed ELP. In the first row, "domain discrepancy": align the features of two domains based on the domain discrepancy; "feature subspace": align the features of two domains in a subspace or via disentanglement; "entropy or metric": enhance the discriminability on the target domain via entropy minimization or metric learning; "self-training on target domain": enhance the discriminability on the target domain via self-supervised learning approach.}
    \label{tab:MainstreamOfDAMethods}
    \begin{tabular}{c|cccc}
        \toprule
        \multirow{2}{*}{Methods} & domain & feature & entropy & self-training on \\
         & discrepancy & subspace & or metric & the target domain \\
        \hline
        SDASL~\cite{yao2015SDASL} &  & $\surd$ &  & \\
        MME~\cite{saito2019semi} &  &  & $\surd$ & \\
        CRST~\cite{zou2019confidence} &  &  & $\surd$ & $\surd$ \\
        dSNE~\cite{xu2019dSNE} & $\surd$ & $\surd$ & $\surd$ & \\
        SHOT~\cite{liang2020shot} &  &  & $\surd$ & $\surd$ \\
        ELP (Ours) & $\surd$ &  & $\surd$ & $\surd$ \\
        \bottomrule
    \end{tabular}
\end{table}

%%%%%
\subsection{Cycle Consistency.} 
Correspondence consistency check by forward-backward mapping between two instances is a widely use technique in image retrieval~\cite{Lowe2004} and stereo matching~\cite{Scharstein2002}. Similar idea was first explored in generative adversarial networks (GANs) by Cycle-GAN~\cite{zhu2017unpaired}, and have been followed up by DA methods such as cycle-consistent adversarial domain adaptation (CyCADA)~\cite{hoffman2018cycada} for domain invariant representation, and temporal cycle consistency learning (TCC)~\cite{dwibedi2019TCC} for the task of temporal alignment between videos.

Different from existing DA methods which try to align source-target feature distributions with adversarial learning techniques or the discrepancy between domains, our method proposes a novel cycle loss to encourage the label consistency after forward-backward mapping between the two domains.
We will demonstrate its effectiveness of promoting feature discriminability in the target domain later.

%%%%%
\subsection{Self-Training.}
Self-training or pseudo-labeling techniques are common practice in semi-supervised learning~\cite{lee2013pseudo,triguero2015self-labeled,saito2017asymmetric,zou2019confidence} and self-supervised learning~\cite{caron2018deep,zhan2020online}.
Typical self-training algorithms generate pseudo labels and use them to retrain the models in an iterative manner. Saito \etal~\cite{saito2017asymmetric} developed an asymmetric self-training scheme that trained two individual classifiers for pseudo labeling and one classifier for the target task.
Zou \etal~\cite{zou2019confidence} introduced the confidence regularized self-training (CRST) that proposed effective regularization strategies to fight against overconfident pseudo labels.
Liang \etal~\cite{liang2020shot} proposed source hypothesis transfer (SHOT) that trained the model on the source domain and then leveraged the centroid for each category in the target domain for pseudo labels to learn a discriminative target-specific feature extractor.

In this paper, we propose a new self-training scheme with memory bank, dynamic threshold strategy, and balanced pseudo-label aggregation.
Different from the existing counterparts, the proposed method is simple to implement and turn out to be effective in eliminating the noise within the pseudo labeled data and facilitating the feature discriminability on the target domain.
We provide an extensive comparison between ELP and representative methods ($\eg$ CRST and SHOT) in Section~\ref{sec:UDAonVisDA-2017}.

% \newpage
\section{Effective Label Propagation Network}

%%%##############################################
\begin{figure*}
    \centering
    \includegraphics[width=0.81\linewidth]{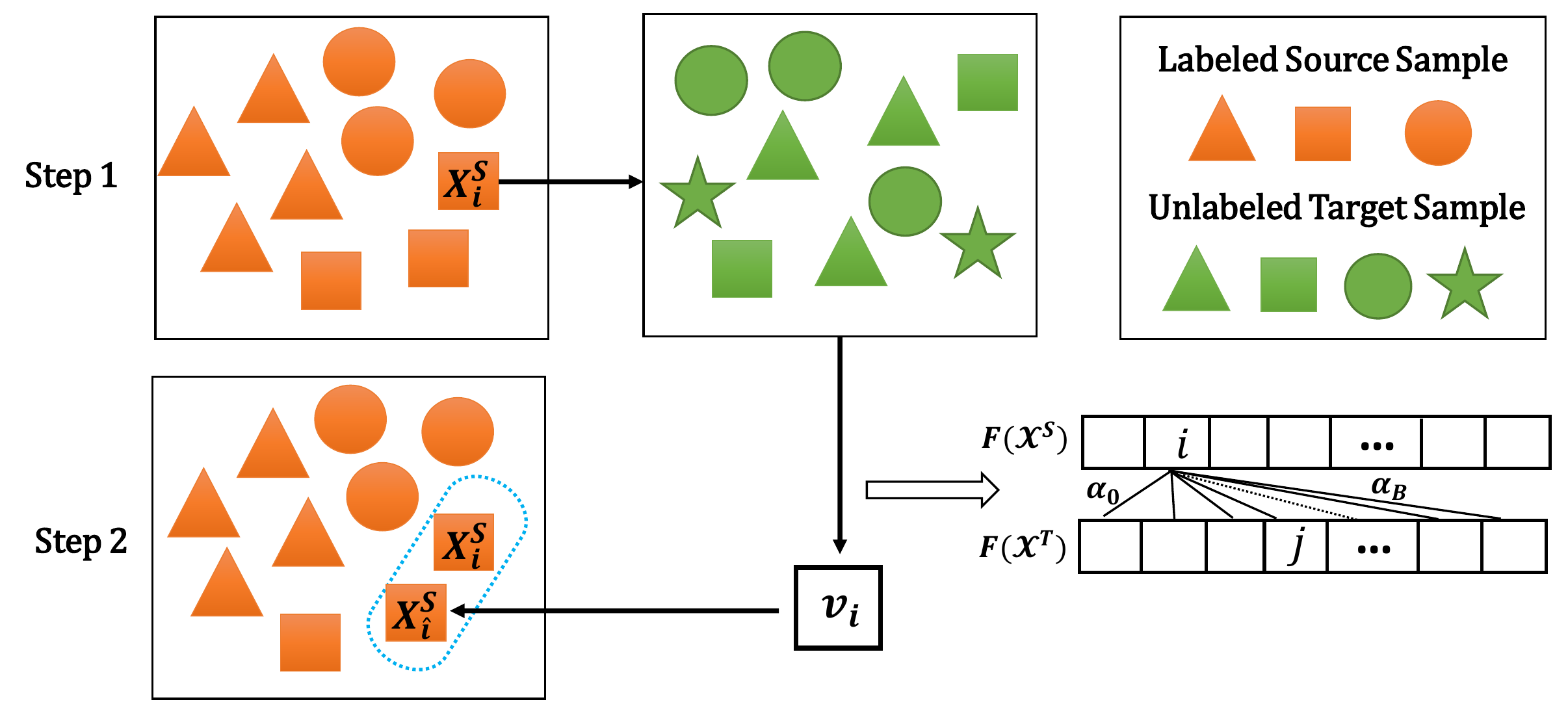}
    \caption{Illustration of the cycle discrepancy loss $\cdl$. For each sample in the source domain, we find its approximation in the target domain and project this approximation back to the source domain.
    Then, we find the corresponding nearest sample of the approximation in the source domain. If the two source samples share the same label, they should stay close in the feature space. $\cdl$ is determined by the distance between such pairs.
    }
    \label{fig:CDL}
\end{figure*}

%%%##############################################
\begin{figure}%[t]
    \centering
    \includegraphics[width=\linewidth]{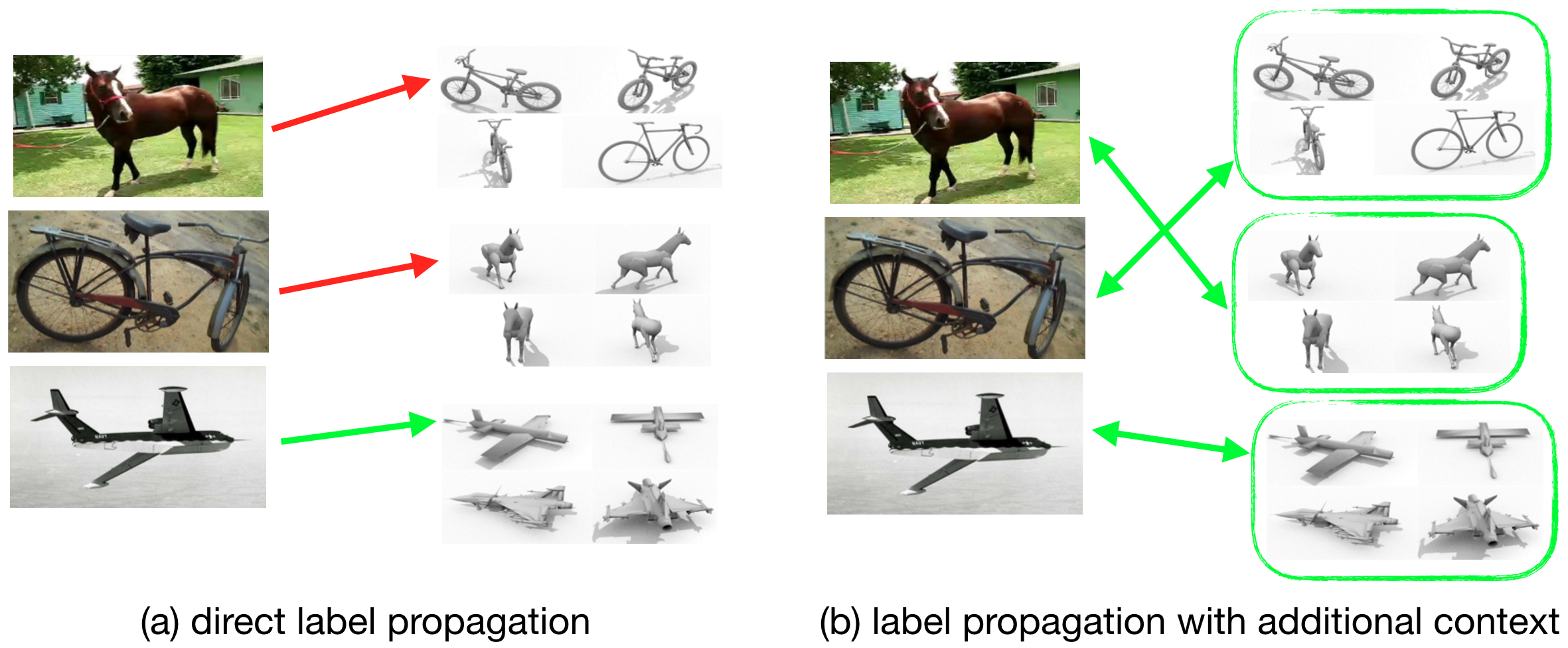}
    \caption{CDL (\textit{right}) helps us strengthen meaningful semantics-aware information propagation between the source domain and the target domain. It is more effective than direct label propagation (\textit{left}).}
    \label{fig:Direct_vs_HighOrder}
\end{figure}

The framework of ELP is illustrated in Fig.~\ref{fig:Framework}: our method extracts features from the labeled and unlabeled samples with a CNN architecture $F$, directs these features to different modules, calculates different losses with a classifier $G$, and minimizes the overall loss in the training process. The key components of ELP include inter-domain propagation (cycle discrepancy loss (CDL)) and intra-domain propagation (a self-training pseudo-labeling scheme with several effective strategies).

We first provide a set of notations for the DA problem in Section~\ref{sec:Preliminary}, and then describe these key components in Section~\ref{sec:CDL} and~\ref{sec:SelfTraining}. Training scheme and implementation details are given in Section~\ref{sec:LearningScheme} and~\ref{sec:Implementation}.

%%%%%%%%%%
\subsection{Preliminary Notations}
\label{sec:Preliminary}
In SSDA, we have data from two visually distinct but semantically related domains, i.e., the source domain $\Sdomain = \{ X^{\Sdomain}, \  Y^{\Sdomain} \}$ where $X^{\Sdomain}, Y^{\Sdomain}$ are data samples and their labels, as well as the target domain $\Tdomain = \{ X^{\Tdomain u} \} \cup \{ X^{\Tdomain l}, \  Y^{\Tdomain l} \}$ where $X^{\Tdomain u}$ are unlabeled samples and $X^{\Tdomain l}, Y^{\Tdomain l}$ are a few labeled samples for each class. Without loss of generality, we describe a DA model with two major components: a feature extractor $F(\cdot)$ and a classifier $G(\cdot)$ which takes the features from $F(\cdot)$ as input and performs the specific inference task. $F(\cdot)$ and $G(\cdot)$ are parameterized by $\mathbf{\theta}_{F}$ and $\mathbf{\theta}_{G}$, respectively.

Basically, the rationale of domain adaptation methodology can be expressed in the following form~\cite{ben2010theory}:
\begin{equation}
        \mathbf{\varepsilon}_{\Tdomain}(h) \leq \mathbf{\varepsilon}_{\Sdomain}(h) + \frac{1}{2} \underbrace{d_{\mathcal{H} \Delta \mathcal{H}}(\Sdomain, \Tdomain)}_{domain \  discrepancy} + \  \gamma,
    \label{equ:RootOfDA}
\end{equation}
where $h$ is a hypothesis of the ideal model, $\mathcal{H}$ is the hypothesis space, $\varepsilon_{\Sdomain}(h), \varepsilon_{\Tdomain}(h)$ are the expected errors of the hypothesis $h$ in the source domain and in the target domain respectively, $d_{\mathcal{H} \Delta \mathcal{H}}(\Sdomain, \Tdomain)$ measures the domain discrepancy with the upper bound of the hypotheses disagreement in $\mathcal{H} \Delta \mathcal{H}$, and $\gamma = {\mathop{min}}_{h} (\varepsilon_{\Tdomain}(h) + \varepsilon_{\Sdomain}(h))$ is the minimum joint hypothesis error. Eq.~(\ref{equ:RootOfDA}) shows that the expected error in the target domain is upper bounded by both $d_{\mathcal{H} \Delta \mathcal{H}}(\Sdomain, \Tdomain)$ and $\gamma$. While most existing DA methods focus on reducing domain discrepancy $d_{\mathcal{H} \Delta \mathcal{H}}(\Sdomain, \Tdomain)$ with better alignment of the feature distributions, our method tries to lower the joint classifications errors $\gamma$ with effective label propagation in the SSDA setting.

%%%%%%%%%%
\subsection{Cycle Discrepancy Loss via Inter-domain Propagation}
\label{sec:CDL}

\subsubsection{Motivation}
% \paragraph{Problem of direct label propagation.}
As discussed previously, most domain adaptation methods attempt to learn a feature extractor for domain alignment, but well aligned features alone do not guarantee good classification performance on the target domain~\cite{chen2019transferability,liang2020shot}.
On the other hand, the direct propagation of the label from the source domain to the target domain is not a good choice.
See Fig.~\ref{fig:Direct_vs_HighOrder} for better understanding of the advantage of high-order propagation over direct label propagation.
Inspired by recent work in cycle consistency~\cite{hoffman2018cycada,dwibedi2019TCC}, we propose a new cycle discrepancy loss (CDL) to encourage samples with the same labels to stay close in the feature space through a cross-domain propagation process.

\subsubsection{Formulation}
Conceptually, CDL is computed for each sample in the source domain, as shown in Fig.~\ref{fig:CDL}: we approximate this sample from the source domain with unlabeled samples in the target domain, project this approximation back into the source domain and locate the nearest source sample. If the two source samples share the same label, we add the CDL loss to penalize the distance between them. In practice, we implement CDL on one mini-batch of labeled samples $B^{\Sdomain}$ from the source domain and one mini-batch of unlabelled samples $B^{\Tdomain u}$ from the target domain. For a sample $\vec{x}_{i}^{\Sdomain} \in B^{\Sdomain}$, we approximate $F(\vec{x}_{i}^{\Sdomain})$ with a new feature vector $\vec{v}_{i}$:
\begin{equation}
    \begin{split}
    \vec{v}_{i} &= \sum_{ \vec{x}_{j}^{
    \Tdomain} \in B^{\Tdomain u} } \alpha_{j} F(\vec{x}_{j}^{
    \Tdomain}), \\
    \alpha_{j} &= \frac{ e^{- || F(\vec{x}_{i}^{\Sdomain}) - F(\vec{x}_{j}^{
    \Tdomain}) ||_{2}^{2}} }{ \sum_{ \vec{x}_{l}^{
    \Tdomain} \in B^{\Tdomain u} } e^{- || F(\vec{x}_{i}^{\Sdomain}) - F(\vec{x}_{l}^{
    \Tdomain}) ||_{2}^{2}} },
    \end{split}
    \label{equ:ProjectFunc}
\end{equation}
where $\vec{v}_i$ is a weighted average of the sample vectors from $B^{\Tdomain u}$, and the weights are determined by the pairwise $L^2$ distance in the feature space. We then use $\vec{v}_i$ as an anchor point in the source domain and calculate a softmax-like score $\beta_j$ for each sample $\vec{x}_{j}^{\Sdomain} \in B^{\Sdomain}$:
\begin{equation}
\beta_{j} = \frac{ e^{- || F(\vec{x}_{j}^{\Sdomain}) - F(\vec{v}_{i}) ||_{2}^{2}} }{ \sum_{ \vec{x}_{l}^{
    \Sdomain} \in B^{\Sdomain} } e^{- || F(\vec{x}_{l}^{\Sdomain}) - F(\vec{v}_{i}) ||_{2}^{2}} }.
    \label{equ:Softmax}
\end{equation}
We denote the sample with the highest score as $\vec{x}_{\hat{i}}^{\Sdomain} = \argmax_{\vec{x}_{j}^{\Sdomain} \in B^{\Sdomain}}\beta_{j}$.
If $\vec{x}_{i}^{\Sdomain}, \vec{x}_{\hat{i}}^{\Sdomain}$ share the same class label, they should be close in the feature space to improve the decision boundaries of this class. The CDL on the mini-batch $B^{\Sdomain}$ is finally formulated as:
\begin{equation}
    \mathcal{L}_{cdl} = - \frac{1}{| B^{\Sdomain} |} \sum_{X_{i}^{\Sdomain} \in B^{\Sdomain}}{\mathop{I}{(y_{i}, y_{\hat{i}})}\log(\beta_{\hat{i}})},
        \label{equ:CDL}
\end{equation}
where $\mathop{I}{(y_{i}, y_{\hat{i}})}$ is a binary indicator function that is activated only when $y_{i}, y_{\hat{i}}$ are equal. As shown in Eqs.~(\ref{equ:ProjectFunc}-\ref{equ:CDL}), minimizing $\mathcal{L}_{cdl}$ will not only encourage the source samples within the same class to stay close to each other, but also drive unlabeled target samples towards nearest labeled source samples.
In this way, the proposed $\mathcal{L}_{cdl}$ helps propagate meaningful semantic-aware information between the two domains.

\subsubsection{Theoretical insight}
To further understand the rationale of $\cdl$, we cast Eqs.~(\ref{equ:CDL}) in another mathematical formulation as follows:
\begin{equation}
\begin{split}
    \cdl &= - \frac{1}{| B^{\Sdomain} |} \sum_{X_{i}^{\Sdomain} \in B^{\Sdomain}}{\yyp\log(\yym)} \\
    & ( \lim_{ \yyp \to 0+ } \yyp \log( \yyp ) = 0, \\
    & \  \lim_{ \yyp \to 1 } \yyp \log( \yyp ) = 0) \\
    &= \frac{1}{| B^{\Sdomain} |} \sum_{X_{i}^{\Sdomain} \in B^{\Sdomain}}[ \yyp\log(\yyp) - \yyp\log(\yym)] \\
    &= \frac{1}{| B^{\Sdomain} |} \sum_{X_{i}^{\Sdomain} \in B^{\Sdomain}}{ \yyp \log( \frac{\yyp}{\yym} ) }
\end{split}
\label{equ:CDL_cast_as_KL}
\end{equation}
It seems that there exists a strong connection between the CDL and the KL-divergence.
Eqs.~(\ref{equ:CDL_cast_as_KL}) indicates that, to minimize $\cdl$ is equal to minimize the KL-divergence from $\yym$ to $\yyp$.
$\yyp$ is the probability distribution based on label information from cycle consistency, %$\cdots$ [+ ASSUMPTION + What the interpolation in (2) implies in the target domain],
and $\yym$ is the probability distribution based on the learned features from cycle consistency or mutual nearest neighbors.
Consequently, the objective of $\cdl$ is to propagate information from $\yyp$ to $\yym$, to learn to model $\yyp$ with $\yym$, and to make $\yym$ discriminative.

%%%%%%%%%%
\subsection{Intra-Domain Propagation via Self Training}
\label{sec:SelfTraining}

Apart from the $\cdl$ that encourages label consistency across the two domains, we also try to improve the feature discriminability on the target domain with an effective self-training strategy.
Our strategy mainly consists of two steps, $\ie$, \emph{aggregation} and \emph{propagation}.
In the aggregation, we record the discriminative information of each class with a set of memory banks ($\eg$, the centroids of the classes in the target domain) and leverage a dynamic threshold strategy to eliminate the noises within the pseudo labels.
In the propagation, we apply a balanced pseudo-label aggregation to strengthen the meaningful training signal in the pseudo-labeled data.

\subsubsection{Aggregation}
We keep a memory bank $\vec{m}_c$ for each class $c$ during the whole training process, which serves as the up-to-date centroid of the class in the target domain. At the start of each epoch, we update $\vec{m}_c$ with all the samples in $X^{\Tdomain l} \cup X^{\Tdomain u}$ that have the label $c$ (denoted as $X_{c}^{\Tdomain}$):
\begin{equation}
    \vec{m}_{c} = \frac{1}{| X_{c}^{\Tdomain} |} \sum_{\vec{x}_{i}^{\Tdomain} \in X_{c}^{\Tdomain}} \frac{F(\vec{x}_{i}^{\Tdomain})}{ || F(\vec{x}_{i}^{\Tdomain}) ||_{2} }.
\end{equation}
Note that we use both unlabeled samples and labeled samples in the target domain for this update.
For unlabeled samples in $X^{\Tdomain u}$, we use the prediction results of the current classifier to determine their pseudo labels.
For a given class $c$, we further use the \textit{median} of the prediction probabilities of all the unlabelled samples as the threshold and select the half above this threshold to update $\vec{m}_c$.
We show in Section~\ref{sec:furtherInvestigation} that this dynamic threshold strategy generally outperforms a static score threshold in refining the pseudo labels. During each iteration of the SGD optimization process, we further smooth $\vec{m}_c$ with the current mini-batch $B^{\Tdomain}$ following~\cite{wu2018unsupervised}:
\begin{equation}
    \vec{m}_{c} = (1 - w) \cdot \vec{m}_{c} + w \cdot \vec{m}_{c}^{B},
    \label{equ:MemoryUpdate}
\end{equation}
where $w$ is a linear weight in $[0, 1]$ and $\vec{m}_{c}^{B}$ is the local memory bank information we collect from this mini-batch. 

\subsubsection{Propagation}
After updating the memory banks for all classes, we can generate pseudo labels for all unlabeled samples in the target domain. For a sample $\vec{x}_{i}^{\Tdomain u} \in X^{\Tdomain u}$, its pseudo label $\hat{y}_{i}$ is determined with the small-loss criteria~\cite{han2018coteaching,jiang2018mentornet}:
\begin{equation}
    \hat{y}_{i} = \argmax_{j} \frac{F(\vec{x}_{i}^{\Tdomain u}) \cdot \vec{m}_{j}}{ || F(\vec{x}_{i}^{\Tdomain u}) ||_{2} }.
    \label{equ:AssignPseudoLabel}
\end{equation}

Rather than using pseudo labels directly for training, we generate a new training sample $(\vec{x}, y)$ by mixing up two pseudo-labeled instances $(\vec{x}_{i}, \hat{y}_{i}), (\vec{x}_{j}, \hat{y}_{j})$ similar to~\cite{zhang2018mixup}:
\begin{equation}
    \begin{split}
        \vec{x} & = (1 - \lambda)\vec{x}_{i} + \lambda \vec{x}_{j}, \\ 
        \hat{y} &=  (1 - \lambda)\hat{y}_{i} + \lambda \hat{y}_{j}, \\
    \end{split}
\end{equation}
where $\lambda$ is a hyper-parameter, and $(\vec{x}_{i}, \hat{y}_{i}), (\vec{x}_{j}, \hat{y}_{j})$ are two samples randomly selected from two mini-batches.
Different from the vanilla mixup in ~\cite{zhang2018mixup} where $\lambda$ randomly varies in $(0, 1)$, we set $\lambda = 0.5$ throughout our experiments. We show in Section~\ref{sec:furtherInvestigation} that this \textit{balanced aggregation} achieves better performance than the vanilla mixup and one recent DA method based on domain mixup~\cite{xu2019adversarial}.
After generating new training samples with the \textit{balanced aggregation}, we calculate the cross entropy loss on these samples and mark it as $\mathcal{L}_{uce}$. 

%%%##############################################
%%%
\begin{algorithm}[t]
    \DontPrintSemicolon
    \KwData{$\{ X^{\mathcal{S}}, Y^{\mathcal{S}} \}, \  \{ X^{\mathcal{T} u} \}, \  \{ X^{\mathcal{T} l}, Y^{\mathcal{T} l} \} $}
    \KwData{$Iter_{1}, \  Iter_{2}, \ w_{cdl}, \ w_{uce}, \ w_{ce}, \ w_{u}$}
    \KwResult{$\mathbf{\theta}_{F}, \  \mathbf{\theta}_{G}$}
    \Begin{
        Initialize the model, iter = 1\;
        \While{$iter \leq Iter_{1}$ {\bf and} $\mathcal{L}$ doesn't converge}{
            iter += 1, randomly sample mini-batches $B^{\mathcal{S}}, \  B^{\mathcal{T} u}, \  B^ {\mathcal{T} l}$\;
            Train the model with $\mathcal{L} = w_{cdl}\mathcal{L}_{cdl} + w_{ce}\mathcal{L}_{ce} + w_{u} \mathcal{L}_{u}$\;
        }
    } % end of Begin of Stage one

    \Begin{
        Get the best model in Stage one\;
        Initialize the memory banks, iter = 1\;
        \While{$iter \leq Iter_{2}$ {\bf and} $\mathcal{L}$ doesn't converge}{
            iter += 1, randomly sample mini-batches $B^{\mathcal{S}}, \  B^{\mathcal{T} u1}, \  B^{\mathcal{T} u2} \  B^ {\mathcal{T} l}$\;
            Calculate pseudo labels for $B^{\mathcal{T} u1}, \  B^{\mathcal{T} u2}$ with the model\;
            Update the memory banks\;
            Refine the pseudo labels via the memory banks\;
            Take $B^{\mathcal{T} u1}$ and $B^{\mathcal{T} u2}$ as input and perform balanced mixup\;
            Train the model with $\mathcal{L} = w_{cdl}\mathcal{L}_{cdl} + w_{uce}\mathcal{L}_{uce} + w_{ce}\mathcal{L}_{ce} + w_{u} \mathcal{L}_{u}$\;
    }
    } % end of Begin of Stage two
    Return $\mathbf{\theta}_{F}, \  \mathbf{\theta}_{G}$
    \caption{The learning scheme of ELP}
    \label{alg:LearningScheme}
\end{algorithm}

% \For{$iter = 1, \cdots, Iter_{1}$}{
%     \If{$condition 1$ {\bf and} $condition 2$}{
%     $Action X_1$\;
%     $Action X_2$\;
%     }
%     \Else{
%     $Action Y_1$\;
%     $Action Y_2$\;
%     }
    
% }

%%%%%%%%%%
\subsection{Overall Learning Scheme}
\label{sec:LearningScheme}
Given one mini-batch of labeled samples $B^{\Sdomain}$ from the source domain, one mini-batch of unlabeled samples $B^{\Tdomain u}$ and one mini-batch of labeled samples $B^{\Tdomain l}$ in the target domain, the total loss $\mathcal{L}$ is a weighted combination of four loss terms:
\begin{equation}
    \mathcal{L} = w_{cdl}\mathcal{L}_{cdl} + w_{uce}\mathcal{L}_{uce} + w_{ce}\mathcal{L}_{ce} + w_{u} \mathcal{L}_{u},
    \label{equ:LossInStageTwo}
\end{equation}
where $\mathcal{L}_{ce}$ is the cross entropy loss for labeled samples in $B^{\Sdomain}$ and $B^{\Tdomain l}$, $\mathcal{L}_{u}$ is the entropy loss for unlabeled samples in $B^{\Tdomain u}$~\cite{saito2019semi}, and $w_{cdl}, w_{uce}, w_{ce}, w_{u}$ are the corresponding weights.
In one practical implementation, ELP works in a two-stage manner, as described in Algorithm~\ref{alg:LearningScheme}: in Stage One, we use $\mathcal{L}_{cdl}, \mathcal{L}_{ce}, \mathcal{L}_{u}$ to transfer semantic information from the source domain to the target domain and learn an initial representation for the target domain.
In Stage Two, we propose $\mathcal{L}_{uce}$ and reinforce the feature discriminability with the self-training scheme.
It should be noted that, the proposed ELP can be easily applied to various domain adaptation methods, including both UDA and SSDA settings, to promote their classification accuracies on the target domain.

%%%%%%%%%%
\subsection{Implementation Details}
\label{sec:Implementation}
\subsubsection{Training}
%%% @ 20200225
Following the training practice in ~\cite{long2018conditional,saito2019semi,lee2019drop}, we use different CNN architectures on different datasets: ResNet-34~\cite{he2016deep} on the DomainNet dataset~\cite{peng2019moment}, VGG-16~\cite{simonyan2014very} on the Office-Home dataset~\cite{venkateswara2017deep}, and ResNet-101~\cite{he2016deep} on the VisDA-2017 dataset~\cite{peng2017visda}. All networks are pre-trained on ImageNet~\cite{deng2009imagenet}, and they serve as feature extractors by removing the last linear classification layer.

To verify the effectiveness of the ELP, we conduct most of our experiments in the SSDA settings and implement ELP based on MME~\cite{saito2019semi}\footnote{From the authors: \url{https://github.com/VisionLearningGroup/SSDA\_MME}.}.
For all the experiments, we train the model using SGD with a momentum value of $0.9$. The initial learning rates for the classifier and the feature extractor are $0.01$ and $0.001$ respectively. We use the learning rate annealing strategy as described in \cite{ganin2015unsupervised}.
We initialize the updating rate of the memory bank $w$ to $0.01$ and increase it linearly with the number of epochs, \emph{i.e.} $w = 0.01 \times epochs$.
The weight parameters $w_{cdl}, w_{uce}, w_{ce}, w_{u}$ are set to $0.1, 0.9, 0.1, 0.1$, respectively, via cross-validation experiments.

To further verify the general effectiveness of the proposed ELP, we also conduct additional experiments in the UDA settings.
In specific, we implement our ELP on DTA~\cite{lee2019drop}\footnote{The official implementation: \url{https://github.com/postBG/DTA.pytorch}.} and analyze whether the ELP can help promote the performance of DTA.
In the UDA experiments, we follow the same configurations as that of DTA. Specifically, we train the model with an initial learning rate of $0.001$, and decay it by a factor of $0.1$ after $10$ training epochs. All networks are trained on one NVIDIA TESLA V100 GPU.

%%%##############################################
\begin{table*}%[!t]
    \centering
    \caption{Results ($\%$) on DomainNet benchmark using
    % VGG-16~\cite{simonyan2014very} and
    ResNet-34~\cite{he2016deep}.
    The table has results of one-shot (1 shot) and three-shot (3 shot) setting on four domains
    For details about number of runs, we mark the first/second with \textbf{bold}/\underline{underline}.
    }
    \label{tab:ResultsOnDomainNet}
    \small{
    \setlength{\tabcolsep}{1.8mm}{
    \begin{tabular}{l|cccccccccccccc|cc}
        \toprule
        \multirow{2}{*}{Method} & \multicolumn{2}{c}{R to C} & \multicolumn{2}{c}{R to P} & \multicolumn{2}{c}{P to C} & \multicolumn{2}{c}{C to S} & \multicolumn{2}{c}{S to P} & \multicolumn{2}{c}{R to S} & \multicolumn{2}{c|}{P to R} & \multicolumn{2}{c}{MEAN} \\
            & \footnotesize{1shot} & \footnotesize{3shot} & \footnotesize{1shot} & \footnotesize{3shot} & \footnotesize{1shot} & \footnotesize{3shot} & \footnotesize{1shot} & \footnotesize{3shot} & \footnotesize{1shot} & \footnotesize{3shot} & \footnotesize{1shot} & \footnotesize{3shot} & \footnotesize{1shot} & \footnotesize{3shot} & \footnotesize{1shot} & \footnotesize{3shot} \\
        % \hline
        % \multicolumn{17}{c}{VGG-16} \\
        % \hline
        % S + T~\cite{ranjan2017l2} & 49.0 & 52.3 & 55.4 & 56.7 & 47.7 & 51.0 & 43.9 & 48.5 & 50.8 & 55.1 & 37.9 & 45.0 & 69.0 & 71.7 & 50.5 & 54.3 \\
        % DANN~\cite{ganin2015unsupervised} & 43.9 & 56.8 & 42.0 & 57.5 & 37.3 & 49.2 & 46.7 & 48.2 & 51.9 & 55.6 & 30.2 & 45.6 & 65.8 & 70.1 & 45.4 & 54.7 \\
        % ADR~\cite{saito2017adversarial} & 48.3 & 50.2 & 54.6 & 56.1 & 47.3 & 51.5 & 44.0 & 49.0 & 50.7 & 53.5 & 38.6 & 44.7 & 67.6 & 70.9 & 50.2 & 53.7 \\
        % CDAN~\cite{long2018conditional} & 57.8 & 58.1 & 57.8 & 59.1 & 51.0 & 57.4 & 42.5 & 47.2 & 51.2 & 54.5 & 42.6 & 49.3 & 71.7 & 74.6 & 53.5 & 57.2 \\
        % ENT~\cite{grandvalet2005semi} & 39.6 & 50.3 & 43.9 & 54.6 & 26.4 & 47.4 & 27.0 & 41.9 & 29.1 & 51.0 & 19.3 & 39.7 & 68.2 & 72.5 & 36.2 & 51.1 \\
        % MME~\cite{saito2019semi} & 60.6 & 64.1 & 63.3 & 63.5 & 57.0 & 60.7 & 50.9 & 55.4 & 60.5 & 60.9 & 50.2 & 54.8 & 72.2 & 75.3 & 59.2 & 62.1 \\
        % \hline
        % MME + ELP & 62.5 & 65.5 & xx.x & xx.x & 59.2 & 63.0 & xx.x & xx.x & 72.8 & 62.1 & xx.x & xx.x & xx.x & 76.0 & xx.x & xx.x \\
        % \hline
        % \hline
        % \multicolumn{17}{c}{ResNet-34} \\
        \hline
        S+T~\cite{ranjan2017l2} & 55.6 & 60.0 & 60.6 & 62.2 & 56.8 & 59.4 & 50.8 & 55.0 & 56.0 & 59.5 & 46.3 & 50.1 & 71.8 & 73.9 & 56.9 & 60.0 \\
        DANN~\cite{ganin2015unsupervised} & 58.2 & 59.8 & 61.4 & 62.8 & 56.3 & 59.6 & 52.8 & 55.4 & 57.4 & 59.9 & 52.2 & 54.9 & 70.3 & 72.2 & 58.4 & 60.7 \\
        ADR~\cite{saito2017adversarial}   & 57.1 & 60.7 & 61.3 & 61.9 & 57.0 & 60.7 & 51.0 & 54.4 & 56.0 & 59.9 & 49.0 & 51.1 & 72.0 & 74.2 & 57.6 & 60.4 \\
        CDAN~\cite{long2018conditional}   & 65.0 & 69.0 & 64.9 & 67.3 & 63.7 & 68.4 & 53.1 & 57.8 & 63.4 & 65.3 & 54.5 & 59.0 & 73.2 & 78.5 & 62.5 & 66.5 \\
        ENT~\cite{grandvalet2005semi} & 65.2 & 71.0 & 65.9 & 69.2 & 65.4 & 71.1 & 54.6 & 60.0 & 59.7 & 62.1 & 52.1 & 61.1 & 75.0 & 78.6 & 62.6 & 67.6 \\
        MME~\cite{saito2019semi} & 70.0 & 72.2 & \underline{67.7} & \underline{69.7} & 69.0 & 71.7 & 56.3 & 61.8 & 64.8 & \underline{66.8} & 61.0 & 61.9 & 76.1 & 78.5 & 66.4 & 68.9 \\
        Meta-MME~\cite{li2020online} & - & 73.5 & - & 70.3 & - & 72.8 & - & 62.8 & - & 68.0 & - & 63.8 & - & 79.2 & - & 70.1 \\
        GVBG~\cite{cui2020GVBG} & \underline{70.8} & \underline{73.3} & 65.9 & 68.7 & \underline{71.1} & \underline{72.9} & \textbf{62.4} & \textbf{65.3} & \underline{65.1} & 66.6 & \textbf{67.1} & \textbf{68.5} & \underline{76.8} & \underline{79.2} & \underline{68.4} & \underline{70.6} \\
        \hline
        MME+Ours & \textbf{72.8} & \textbf{74.9} & \textbf{70.8} & \textbf{72.1} & \textbf{72.0} & \textbf{74.4} & \underline{59.6} & \underline{64.3} & \textbf{66.7} & \textbf{69.7} & \underline{63.3} & \underline{64.9} & \textbf{77.8} & \textbf{81.0} & \textbf{69.0} & \textbf{71.6} \\
 %                                & +2.8 & +2.7 & +3.1 & +2.4 & +3.0 & +2.7 & +3.3 & +2.5 & +1.9 & +2.9 & +2.3 & +3.0 & +1.7 & +2.4 & +2.6 & +2.7 \\
        \bottomrule
    \end{tabular}
    }
    \vspace{-3mm}
    }
\end{table*}

\subsubsection{Inference}
In the inference stage, we resize the shortest edge of input images to $256$, crop $224 \times 224$ patches from the center part, and report the classification accuracy based on the cropped patches, which is consistent with previous works~\cite{long2018conditional,saito2019semi,saito2018MCD,lee2019drop}.
For each adaptation scenario, we repeat the experiments three times and report the average accuracy.
\section{Experiments}
\label{sec:Experiments}
In this section, we evaluate the performance of our ELP on multiple domain adaptation benchmarks and compare the results with several state-of-the-art DA approaches.

%%%%%%%%%%%%%%%%%%%%%
\subsection{Datasets}
We validate our method on three public DA benchmarks.
For fair comparison, we randomly run our method for three times with different random seeds via Pytorch~\cite{paszke2019pytorch} and report the average classification accuracies.

\textbf{Office-Home}~\cite{venkateswara2017deep} contains approximately $15,500$ images from $4$ image domains ($\ie$ \textbf{R}eal, \textbf{C}lipart, \textbf{A}rt, and \textbf{P}roduct) with $65$ classes.
Follow the common practice, we conduct comparison experiments on $12$ adaptation scenarios in total.

\textbf{DomainNet}~\cite{peng2019moment} contains 6 domains with 345 classes.
It is usually used for testing large-scale domain adaptation.
However, labels of some domains and classes are very noisy.
Thus, by following the experimental configuration of \cite{saito2019semi}, $4$ domains (\textbf{R}eal, \textbf{C}lipart, \textbf{P}ainting, and \textbf{S}ketch) and $126$ classes are selected.
We conduct comparison experiments on the adaption scenarios where the target domains are different from real by picking up 7 scenarios from the domains, following the same configurations in MME~\cite{saito2019semi}.

%%%##############################################
\begin{figure}
    \centering
    \includegraphics[width=\linewidth]{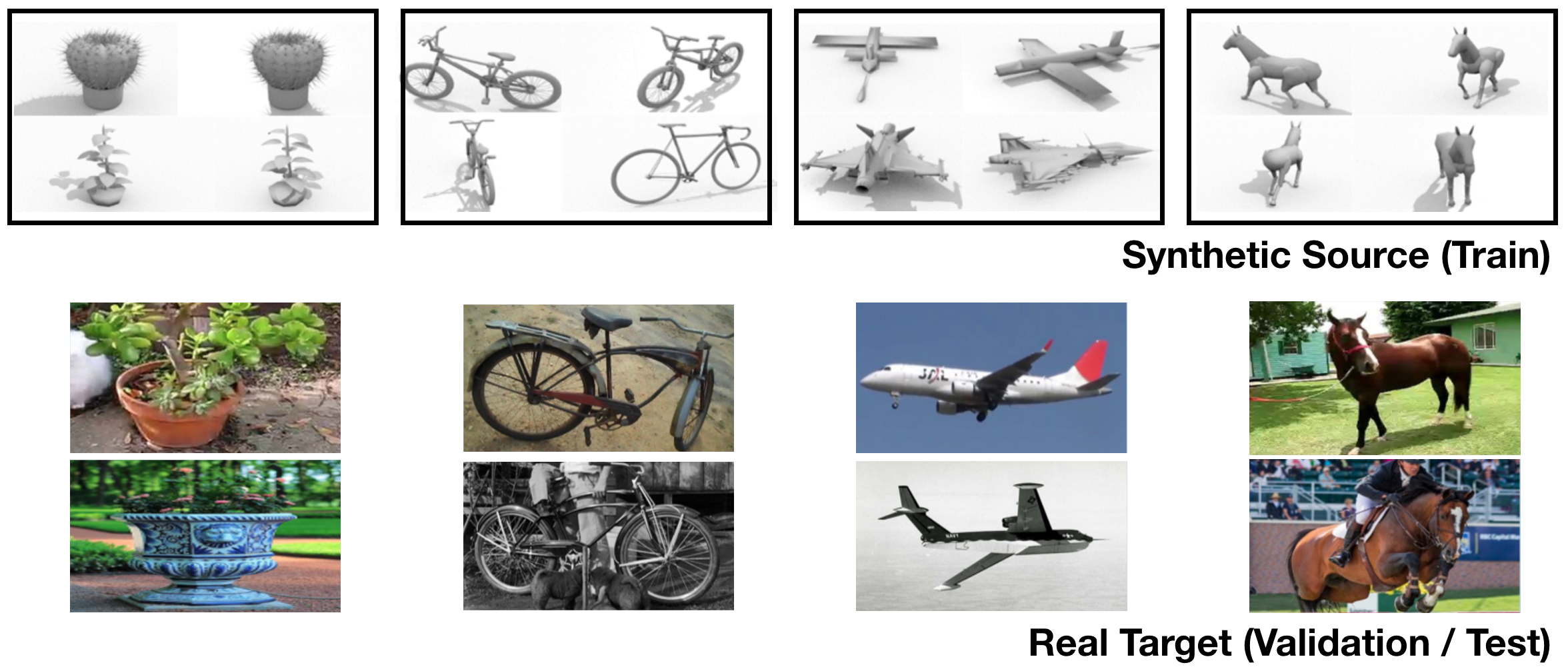}
    \caption{Some instances from the synthetic domain (source) and the real (target) domains in the VisDA-2017 benchmark~\cite{peng2017visda}.
    }
    \label{fig:VisDA2017Instances}
    \vspace{-3mm}
\end{figure}

\textbf{VisDA-2017}~\cite{peng2017visda} contains approximately $280,000$ images from $12$ classes (see the first row in Table~\ref{tab:ResultsOnVisDA2017}).
It is a challenging benchmark for testing the domain shift from \emph{synthetic} data to \emph{real} images with complex background components (see Fig.~\ref{fig:VisDA2017Instances} for instance).
The synthetic data are renderings of 3D models from different angles and with different lighting conditions.
VisDA-2017 contains two tasks: classification and segmentation.
In our experiments, we focus on the classification task.
Follow the common practices~\cite{peng2017visda,saito2018maximum,zou2019confidence,lee2019drop}, we split the images into three sets: a training set with $152,397$ synthetic images, a validation set with $55,388$ real-world images, and a test set with $72,372$ real-world images. We adopt the synthetic images as source domain and the validation set as target domain, which means that we report the classification performance on the validation partition.

%%%##############################################
\begin{table*}
    \centering
    \caption{Comparisons of different approaches on the three-shot SSDA scenarios from Office-Home with ResNet-50.}
    \label{tab:ResultsOnOfficeHome_ResNet50}
    \footnotesize{
    \setlength{\tabcolsep}{3.6mm}{
    \begin{tabular}{l|cccccccc|c}
        \toprule
        Method & A $\to$ C & A $\to$ P & A $\to$ R & C $\to$ A & C $\to$ R & C $\to$ P & P $\to$ A & P $\to$ C & MEAN \\
        \hline
        \multicolumn{10}{c}{Three-shot SSDA} \\
        \hline
        MME~\cite{saito2019semi} & 60.5 & 77.4 & 76.3 & 64.0 & 73.4 & 73.7 & 63.3 & 61.6 & 68.8 \\
        MME + Ours & 62.0 & 78.3 & 77.5 & 64.9 & 75.3 & 79.0 & 64.4 & 63.1 & 70.6 \\
        \bottomrule
    \end{tabular}
    }
    }
\end{table*}
%%%##############################################
\begin{table*}
    \centering
    \caption{Comparisons of different methods on Office-Home dataset based on VGG-16.
    The results ($\%$) include one-shot and three-shot on all possible adaptation scenarios. For details about number of runs, we mark the first / second with \textbf{bold} / \underline{underline}.
    }
    \label{tab:ResultsOnOfficeHome}
    \small{
    \setlength{\tabcolsep}{4.1mm}{
    \begin{tabular}{l|c|c|c|c|c}
        \toprule
        Method & R / P / A $\to$ C & R / A / C $\to$ P & R / P / C $\to$ A & P / A / C $\to$ R & MEAN \\
        % \hline
        % \multicolumn{6}{c}{UDA on ResNet-50 (results from ~\cite{chen2019transferability})} \\
        % \hline
        % DANN~\cite{ganin2015unsupervised} & 51.8 / 43.7 / 45.6 & 76.8 / 59.3 / 58.5 & 63.2 / 46.1 / 47.0 & 68.5 / 70.1 / 60.9 & 57.6 \\
        % CDAN~\cite{long2018conditional} & 56.7 / 50.9 / 50.7 & 81.6 / 70.6 / 70.0 & 70.9 / 57.4 / 57.6 & 77.3 / 76.0 / 70.0 & 65.8 \\
        % BSP~\cite{chen2019transferability} & 59.3 / 50.2 / 52.0 & 81.9 / 68.6 / 70.3 & 72.2 / 58.6 / 58.0 & 77.6 / 76.1 / 70.2 & 66.3 \\
        \hline
        \multicolumn{6}{c}{One-shot SSDA} \\
        \hline
        S + T~\cite{ranjan2017l2} &  39.5 / 37.0 / 37.5  & 75.3 / 63.6 / 65.9 & 61.2 / 52.0 / 51.4 & 71.6 / 69.5 / 64.5 & 57.4 \\
        DANN~\cite{ganin2015unsupervised} &  \textbf{52.0} / 45.9 / 44.4  & 75.7 / 64.3 / 65.3 & 62.7 / 51.3 / 52.3 & 72.7 / 68.9 / 64.2 & 60.0 \\
        ADR~\cite{saito2017adversarial}   &  39.7 / 37.2 / 39.0  & 76.2 / 63.9 / 65.2 & 60.2 / 51.4 / 50.0 & 71.8 / 68.7 / 64.8 & 57.4 \\
        CDAN~\cite{long2018conditional}   &  43.3 / 37.4 / 39.8  & 75.7 / 67.7 / 66.2 & 60.9 / 44.5 / 41.6 & 69.6 / 64.8 / 58.7 & 55.8 \\
        ENT~\cite{grandvalet2005semi}     &  23.7 / 21.3 / 22.4  & 77.5 / 66.0 / 67.7 & 64.0 / 44.6 / 25.1 & \underline{74.6} / 70.6 / 62.1 & 51.6 \\
        MME~\cite{saito2019semi}          &  49.1 / \underline{46.2} / \underline{45.8}  & \underline{78.7} / \underline{68.6} / \underline{71.3} & \underline{65.1} / \underline{56.0} / \textbf{57.5} & 74.4 / \underline{72.2} / \underline{68.0} & \underline{62.7} \\
        \hline
        MME + Ours & \underline{49.2} / \textbf{46.7} / \textbf{46.1} & \textbf{79.7} / \textbf{69.0} / \textbf{71.6} & \textbf{65.5} / \textbf{56.3} / \underline{57.4} & \textbf{75.3} / \textbf{72.4} / \textbf{68.2} & \textbf{63.1} \\
        \hline
        \hline
        \multicolumn{6}{c}{Three-shot SSDA} \\
        \hline
        S + T~\cite{ranjan2017l2}         &  49.6 / 47.2 / 47.5  & 78.6 / 69.4 / 70.4 & 63.6 / 55.9 / 56.2 & 72.7 / 73.4 / 69.7 & 62.9 \\
        DANN~\cite{ganin2015unsupervised} &  56.1 / 52.4 / 50.0  & 77.9 / 69.5 / 69.8 & 63.7 / 56.3 / 56.4 & 73.6 / 72.3 / 68.7 & 63.9 \\
        ADR~\cite{saito2017adversarial}   &  49.0 / 47.8 / 49.3  & 78.1 / 69.9 / 71.4 & 62.8 / 55.8 / 45.3 & 73.6 / 73.3 / 69.3 & 63.0 \\
        CDAN~\cite{long2018conditional}   &  50.2 / 45.1 / 46.0  & 80.9 / 74.7 / 71.2 & 62.1 / 50.3 / 52.9 & 70.8 / 71.4 / 65.9 & 61.8 \\
        ENT~\cite{grandvalet2005semi}     &  48.3 / 46.8 / 44.8  & 81.6 / 73.0 / \textbf{77.0} & 65.5 / 56.9 / 59.1 & \underline{76.6} / \underline{75.3} / 72.9 & 64.8 \\
        MME~\cite{saito2019semi}          &  \underline{56.9} / \underline{53.6} / \underline{54.9}  & \underline{82.9} / \underline{75.7} / \underline{76.3} & \underline{65.7} / \underline{59.2} / \underline{61.1} & \textbf{76.7} / \underline{75.3} / \underline{72.9} & \underline{67.6} \\
        \hline
        MME + Ours & \textbf{57.1} / \textbf{53.9} / \textbf{55.1} & \textbf{83.2} / \textbf{75.9} / 76.1 & \textbf{67.0} / \textbf{59.3} / \textbf{61.9} & 76.3 / \textbf{76.3} / \textbf{73.3} & \textbf{68.0} \\
        \bottomrule
    \end{tabular}
    }
    \vspace{-3mm}
    }
\end{table*}

%%%%%%%%%%%%%%%%%%%%%
\subsection{Baselines} We choose multiple \textit{state-of-the-art} DA approaches, including both typical methods and latest ones, as the baselines:
S + T~\cite{ranjan2017l2}, % "is a model trained with the labeled source and labeled target examples without using unlabeled target examples."
entropy minimization (ENT)~\cite{grandvalet2005semi}, %"is a model trained with labeled source and target and unlabeled target using entropy minimization.
domain adversarial neural network (DANN)~\cite{ganin2015unsupervised}, %"employs a domain classifier to match feature distributions. %This is the most popular method of UDA.
%Entropy is calculated on unlabeled target examples and entire network is trained to minimize it.
domain adaptation network (DAN)~\cite{long2015learning},
adversarial dropout regularization (ADR)~\cite{saito2017adversarial}, %utilizes a task-specific decision boundary to align features. This method is proposed to extract discriminative target features.
conditional domain adaptation network (CDAN)~\cite{long2018conditional}, %performs conditional domain alignment on features based on the output of classifiers. %In addition, it utilizes entropy minimization on target instances.  CDAN integrates domain-classifier based alignment and entropy minimization.
maximum classifier discrepancy (MCD)~\cite{saito2018MCD},
CRST~\cite{zou2019confidence}, %""
batch spectral penalization (BSP)~\cite{chen2019transferability}, %"penalizes the largest singular values to promote the feature discriminability on the target domain."
adaptive feature normalization (AFN)~\cite{xu2019AFN}, %"$\cdots$"
dSNE~\cite{xu2019dSNE}, drop to adapt (DTA)~\cite{lee2019drop}, %"leverages adversarial dropout to learn strongly discriminative features by enforcing the cluster assumption."
MME~\cite{saito2019semi}, adversarial domain adaptation with domain mixup (DM-ADA)~\cite{xu2019adversarial}, gradually vanishing bridge (GVBG)~\cite{cui2020GVBG}, meta-learning based framework (Meta-)~\cite{li2020online}, and SHOT~\cite{liang2020shot}.

We will demonstrate the effective of our method in improving classification accuracies of the aforementioned baselines on the target domain.
For reliable results, we cited the performance of these baseline methods from ~\cite{chen2019transferability,lee2019drop,saito2019semi}.

%%%%%%%%%%%
\subsection{SSDA Results on DomainNet.}
We demonstrate the effectiveness of our ELP by comparing it with several existing counterparts on DomainNet benchmark.
We apply %VGG-16~\cite{simonyan2014very} and
ResNet-34~\cite{he2016deep}, the same backbone network in the previous work~\cite{saito2018MCD,long2018conditional,chen2019transferability,lee2019drop}.
For the results of baselines, we directly reported their values from the original paper~\cite{saito2019semi}\footnote{For the results of Meta-MME~\cite{li2020online}, the authors only reported their three-shot SSDA accuracies in their paper.}.

The classification accuracy per adaptation setting and the total average results on the DomainNet dataset are listed in Table~\ref{tab:ResultsOnDomainNet}.
We can observe that, among all the methods, the proposed ELP achieves the best classification accuracy of $69.0\% \  (+2.6\%)$ in one-shot SSDA and $71.5\% \ (+2.7\%)$ in three-shot SSDA, respectively.
Besides, on all the $7 \times 2$ adaptation settings, our ELP consistently improves the classification accuracy on MME.
Moreover, to demonstrate the versatility of our ELP, we randomly select $7$ adaptation settings from Table~\ref{tab:ResultsOnDomainNet} and conduct similar comparison experiments using VGG-16~\cite{simonyan2014very}. The results indicates that on one different network backbone, the proposed ELP can also boost the classification accuracies of MME by $1 \sim 3 \%$.
% The superiority of our method further demonstrates that the improved feature discriminability on the target domain and verifies the effectiveness of the proposed ELP.

%\textcolor{green}{Plus some sentences to analyze the reason why ELP works. Also be careful whether there exists some weird circumstance}

%%%##############################################
\begin{table*}
    \centering
    \caption{Classification accuracy ($\%$) on the VisDA-2017 validation partition based on ResNet-101. S-only means training the model with the source domain alone.
    For details about number of runs, we mark the first/second with \textbf{bold}/\underline{underline}.
    }
    \label{tab:ResultsOnVisDA2017}
    \small{
    \setlength{\tabcolsep}{2.8mm}{
    \begin{tabular}{l|cccccccccccc|c}
        \toprule
        Method & \rotatebox{90}{airplane} & \rotatebox{90}{bicycle} & \rotatebox{90}{bus} & \rotatebox{90}{car} & \rotatebox{90}{horse} & \rotatebox{90}{knife} & \rotatebox{90}{motorcycle} & \rotatebox{90}{person} & \rotatebox{90}{plant} & \rotatebox{90}{skateboard} & \rotatebox{90}{train} & \rotatebox{90}{truck} & MEAN \\
        % \hline
        % \multicolumn{14}{c}{Unsupervised Domain Adaptation} \\
        \hline
        % S-only & 46.2 & 27.6 & 31.4 & 78.1 & 71.8 & 1.4 & 71.6 & 14.3 & 63.5 & 31.0 & \textbf{93.7} & 3.2 & 50.8 \\
        S-only & 55.1 & 53.3 & 61.9 & 59.1 & 80.6 & 17.9 & 79.7 & 31.2 & 81.0 & 26.5 & 73.5 & 8.5 & 52.4 \\
        % S-finetune & 72.3 & 6.1 & 63.4 & 91.7 & 52.7 & 7.9 & 80.1 & 5.6 & 90.1 & 18.5 & 78.1 & 25.9 & 49.4 \\
        % RevGrad~\cite{ganin2015unsupervised} & 81.9 & 77.7 & 82.8 & 44.3 & 81.2 & 29.5 & 65.1 & 28.6 & 51.9 & 54.6 & 82.8 & 7.8 & 57.4 \\
        DAN~\cite{long2015learning}  & 68.1 & 15.4 & 76.5 & \textbf{87.0} & 71.1 & 48.9 & 82.3 & 51.5 & 88.7 & 33.2 & \underline{88.9} & 42.2 & 62.8 \\
        MCD~\cite{saito2018maximum} & 87.0 & 60.9 & 83.7 & 64.0 & 88.9 & 79.6 & 84.7 & 76.9 & 88.6 & 40.3 & 83.0 & 25.8 & 71.9 \\
        ADR~\cite{saito2017adversarial} & 87.8 & 79.5 & 83.7 & 65.3 & 92.3 & 61.8 & 88.9 & 73.2 & 87.8 & 60.0 & 85.5 & 32.3 & 74.8 \\
        %% remove the follow two counterparts @ 20200106
        % SE~\cite{french2018SE}           & 95.9 & 87.4 & 85.2 & 58.6 & 96.2 & 95.7 & 90.6 & 80.0 & 94.8 & 90.8 & 88.4 & 47.9 & 84.3 \\
        %CAN          & 97.0 & 87.2 & 82.5 & 74.3 & 97.8 & 96.2 & 90.8 & 80.7 & 96.6 & 96.3 & 87.5 & 59.9 & 87.2 \\
        %% add the following two counterparts @ 20200106
        CDAN~\cite{long2018conditional}   & 85.2 & 66.9 & 83.0 & 50.8 & 84.2 & 74.9 & 88.1 & 74.5 & 83.4 & 76.0 & 81.9 & 38.0 & 73.7 \\
        DM-ADA~\cite{xu2019adversarial} & - & - & - & - & - & - & - & - & - & - & - & - & 75.6 \\
        CDAN+BSP~\cite{chen2019transferability} & 92.4 & 61.0 & 81.0 & 57.5 & 89.0 & 80.6 & 90.1 & 77.0 & 84.2 & 77.9 & 82.1 & 38.4 & 75.9 \\
        AFN~\cite{xu2019AFN} & 93.6 & 61.3 & \underline{84.1} & 70.6 & \textbf{94.1} & 79.0 & \textbf{91.8} & 79.6 & 89.9 & 55.6 & \textbf{89.0} & 24.4 & 76.1 \\
        CRST~\cite{zou2019confidence} & 88.0 & 79.2 & 61.0 & 60.0 & 87.5 & \underline{81.4} & 86.3 & 78.8 & 85.6 & \textbf{86.6} & 73.9 & \underline{68.8} & 78.1 \\
        SHOT~\cite{liang2020shot} & 92.6 & 81.1 & 80.1 & 58.5 & 89.7 & \textbf{86.1} & 81.5 & 77.8 & 89.5 & 84.9 & 84.3 & 49.3 & 79.6 \\
        dSNE~\cite{xu2019dSNE} & - & - & - & - & - & - & - & - & - & - & - & - & 80.7 \\
        DTA~\cite{lee2019drop} & \underline{93.7} & \underline{82.2} & \textbf{85.6} & \underline{83.8} & 93.0 & 81.0 & \underline{90.7} & \underline{82.1} & \textbf{95.1} & 78.1 & 86.4 & 32.1 & \underline{81.5} \\
        \hline
        DTA + Ours & \textbf{98.0} & \textbf{88.4} & 74.3 & 76.0 & \underline{93.6} & 81.1 & 87.5 & \textbf{83.1} & \underline{93.3} & \underline{86.0} & 87.3 & \textbf{84.4} & \textbf{86.1} \\
        % \hline
        % \multicolumn{14}{c}{Semi-Supervised Domain Adaptation (three-shot)} \\
        % \hline
        % %ResNet-101~\cite{he2016deep} &  &  &  &  &  &  &  &  &  &  &  &  &  \\
        % %baselineX    &  &  &  &  &  &  &  &  &  &  &  &  &  \\
        % MME~\cite{saito2019semi} & 90.6 & 83.0 & 63.7 & 74.5 & 91.5 & 48.1 & 78.7 & 71.8 & 92.0 & 82.5 & 76.6 & 30.1 & 73.59 \\
        % MME~\cite{saito2019semi} + Ours & 97.9 & 86.9 & 73.2 & 78.2 & 92.7 & 49.1 & 80.3 & 68.0 & 94.7 & 88.3 & 77.4 & 39.5 & 77.18 \\
        \bottomrule
    \end{tabular}
    }
    % \vspace{-3mm}
    }
\end{table*}

%%%%%%%%%%%
\subsection{SSDA Results on Office-Home.}
We also compare the proposed ELP with some state-of-the-art methods, including CDAN~\cite{long2018conditional}, ENT~\cite{grandvalet2005semi}, and MME~\cite{saito2019semi}, on Office-Home benchmark.
We apply VGG-16~\cite{simonyan2014very} and ResNet-50~\cite{he2016deep} as the backbone for fair comparisons.

The results are listed in Table~\ref{tab:ResultsOnOfficeHome_ResNet50} and Table~\ref{tab:ResultsOnOfficeHome}.
As it can be observed that, our method achieves the best classification accuracy of $68.0\%$ on average (three-shot SSDA with VGG-16), better than other SSDA counterparts.
And on ResNet-50, the proposed ELP also helps improve the classification performance of MME.
We also notice that:
(1) asymmetric transferability generally exists, like $75.9\%$ (Art $\to$ Product) $\neq 59.3\%$ (Product $\to$ Art), and $73.3\%$ (Clipart $\to$ Real) $\neq 57.1\%$ (Real $\to$ Clipart). It necessitates the refinement of label propagation procedure between the source domain and the target domain.
(2) For each sub-dataset pairs, the large amount of the source domain generally leads to the degradation in the classification accuracy on the target domain. These observations indicate the indispensable role of label information propagation in the target domain for improved feature discriminability and better classification performance.
% (x) There are still some adaptation settings that UDA methods outperform SSDA models, like (Real $\to$ Art). It calls for further developments of SSDA methods.

%%%%%%%
\subsection{UDA results on VisDA-2017.}
\label{sec:UDAonVisDA-2017}
To extensively validate the versatility of our method in DA, we investigate whether the proposed ELP can help achieve better results in UDA setting. %, where no manual annotation of the target domain is available.
Without loss of generality, we choose DTA~\cite{lee2019drop} as the baseline and boost it with the proposed ELP on ResNet-101~\cite{he2016deep}, the same backbone in ~\cite{long2018conditional,saito2018MCD,lee2019drop,long2015learning}, on the VisDA-2017 dataset, and report the performance on the validation partition.

\begin{figure*}
    \centering{
    \includegraphics[width=\linewidth]{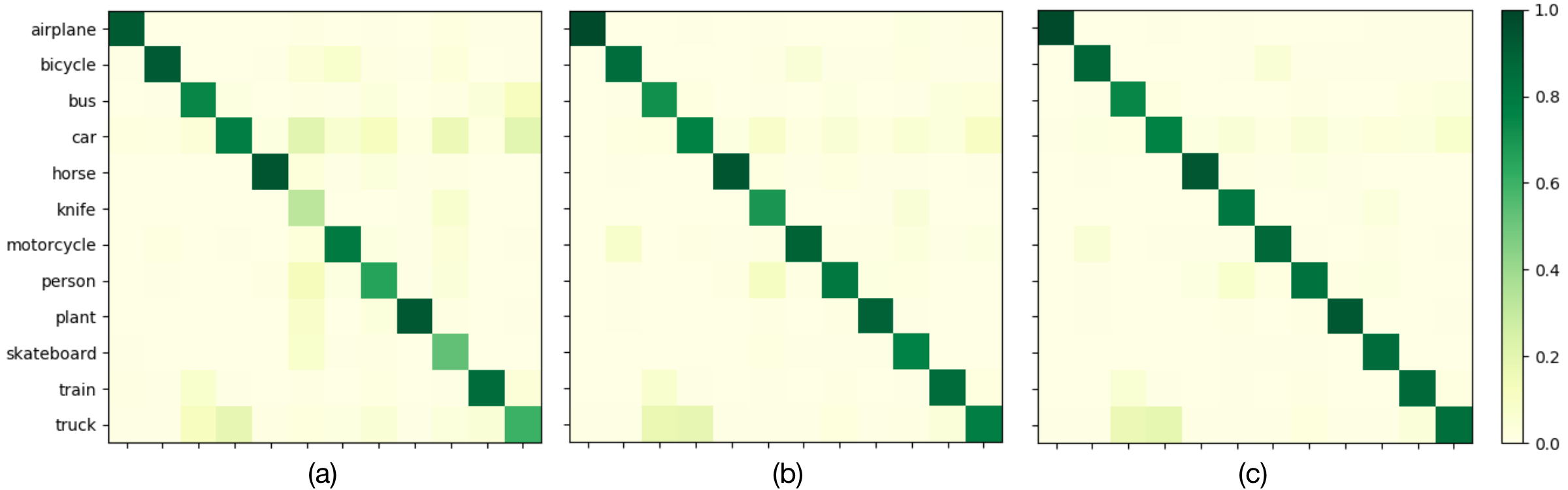}
    \caption{\textcolor{black}{The visualizations of the confusion matrix from different methods on VisDA-2017 dataset. (a) BSP~\cite{chen2019transferability} (b) DTA~\cite{lee2019drop} (c) DTA + ours}.}
    \label{fig:ConfusionMatrixOnVisDA2017}
    }
\end{figure*}

\begin{figure*}
    \centering
    \includegraphics[width=\linewidth]{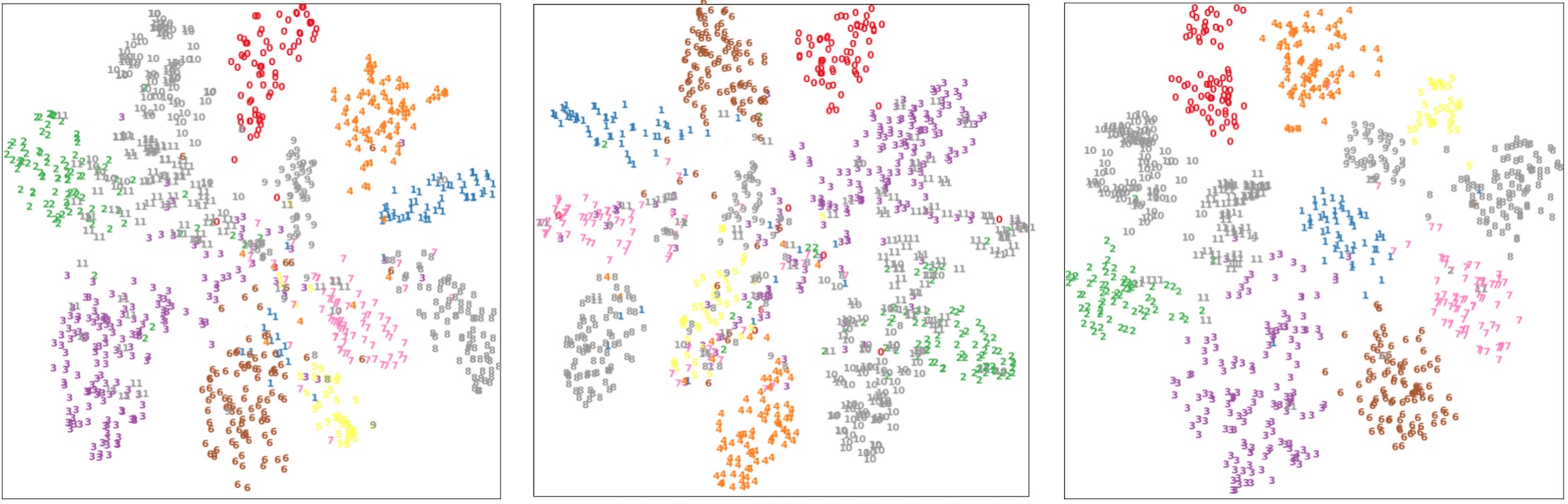}
    \caption{A t-SNE visualization~\cite{maaten2008tsne} of the embedded features for the \emph{synthetic-to-real} task from the VisDA-2017 dataset~\cite{peng2017visda}. Features from classes $0$ to $11$ are marked with different colors. (\textit{Left}) MME~\cite{saito2019semi}, (\textit{Middle}) DTA~\cite{lee2019drop}, and (Right) DTA + the proposed ELP.
    }
    \label{fig:MMEvsDTAvsOurs}
\end{figure*}

The detailed comparisons are listed in Table~\ref{tab:ResultsOnVisDA2017}\footnote{For the results of dSNE~\cite{xu2019dSNE} and DM-ADA~\cite{xu2019adversarial}, since the authors just reported the mean accuracy, we cannot get its performance on each class.}.
As we can observe that DTA + our ELP achieves the best classification performance $86.1\%$ and gains $4.6\%$ better than DTA~\cite{lee2019drop}, in terms of mean average per class.
It outperforms other baselines that apply self-supervised pseudo-labeling strategy, such as CRST~\cite{zou2019confidence}, dSNE~\cite{xu2019dSNE}, and SHOT~\cite{liang2020shot}.
Confusion matrix in Fig.~\ref{fig:ConfusionMatrixOnVisDA2017} and feature visualization in Fig.~\ref{fig:MMEvsDTAvsOurs} also help demonstrate the feature discriminability on the target domain ameliorated by the proposed method.
Thus, it is safe to say that applying our ELP generally helps cultivate better feature discriminability and boost classification results in domain adaptation.

% \textcolor{green}{Why the margin on \textit{truck} class is so big? Visualize the confusion matrix of DTA and DTA + ELP for further analyses (see Fig~\ref{fig:ConfusionMatrixOnVisDA2017}).
% What if our paper is reviewed by the authors of CAN or what if the reviewer knows CAN?
% Do we need to conduct more experiments (\emph{e.g.} CDAN~\cite{long2018conditional} + ELP, SE~\cite{french2018SE} + ELP, or ELP alone, or using other network to compare with other baseline ?) to verify the power of ELP?}

%%%##############################################
%%%##############################################
%%%##############################################
\begin{figure*}
    \centering
    \includegraphics[width=\linewidth]{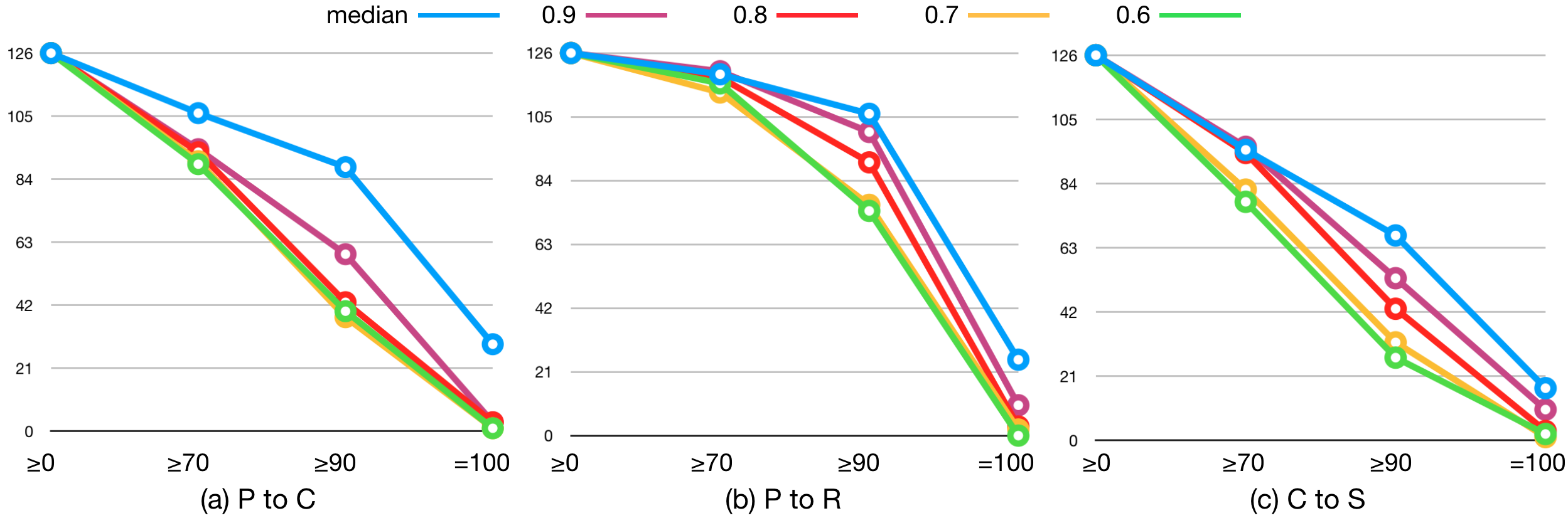}
    \caption{Comparisons of using median and different fixed thresholds to separate the clean data and the noisy data. The results are generated by conducting three adaptation tasks on the DomainNet dataset. The horizontal axis represents different accuracy of the clean data, and the vertical axis represents the total number of categories in the corresponding interval.}
    \label{fig:MedianVSOthers}
\end{figure*}

%%%%%%%%%%%%%%%%%%%%%%%%%%%%
\begin{table*}
    \centering
    \caption{Ablation study of the two stage learning scheme on the DomainNet dataset.}
    \label{tab:AblationStudyOfTwoStageScheme}
    \small{
    \begin{tabular}{l|ccccccc|c}
        \toprule
        Setting & P $\to$ R & R $\to$ C & R $\to$ S & R $\to$ P & P $\to$ C & C $\to$ S & S $\to$ P & MEAN\\
        \hline
            \multicolumn{9}{c}{Three-Shot SSDA} \\
        \hline
        MME~\cite{saito2019semi} & 78.5 & 72.2 & 61.9 & 69.7 & 71.7 & 61.8 & 66.8 & 68.9 \\ 
        w/o stage two & 78.3 & 72.8 & 63.1 & 69.8 & 71.8 & 62.6 & 67.3 & 69.4 \\ 
        Full model (ELP)  & 81.0 & 74.1 & 64.9 & 72.1 & 74.4 & 64.4 & 69.7 & 71.5 \\ 
        \bottomrule
    \end{tabular}
    }
\end{table*}

\begin{table*}
    \centering
    \caption{Performance variation caused by the hyper-parameter $\lambda$ in the proposed intra-domain label propagation. The comparison experiments are conducted on the DomainNet dataset and ResNet-34.}
    \label{tab:AblationStudyTheAlphaInMixUp}
    \small{
    \begin{tabular}{l|ccccccc|c}
        \toprule
        Setting & P $\to$ R & R $\to$ C & R $\to$ S & R $\to$ P & P $\to$ C & C $\to$ S & S $\to$ P & MEAN\\
        \hline
        \multicolumn{9}{c}{Three-Shot SSDA} \\
        \hline
        MME & 78.5 & 72.2 & 61.9 & 69.7 & 71.7 & 61.8 & 66.8 & 68.9 \\
        w/o stage two & 78.2 & 72.8 & 63.1 & 69.8 & 71.8 & 62.6 & 67.3 & 69.4 \\
        \hline
        random $\lambda$ & 79.7 & 73.6 & 63.3 & 70.8 & 72.4 & 62.6 & 67.9 & 70.0 \\
        $\lambda= 0$ & 78.9 & 72.9 & 62.6 & 70.3 & 71.9 & 62.3 & 67.3 & 69.5 \\
        $\lambda= 0.1$ & 79.2 & 73.1 & 64.3 & 70.5 & 72.1 & 62.2 & 67.3 & 69.8\\
        $\lambda= 0.2$ & 79.9 & 73.5 & 64.6 & 70.8 & 72.7 & 62.7 & 67.9 & 70.3 \\
        $\lambda= 0.3$ & 80.6 & 73.4 & 64.7 & 71.4 & 73.4 & 63.3 & 69.0 & 70.8 \\
        $\lambda= 0.4$ & 80.8 & 74.0 & 64.7 & 71.3 & 73.9 & 64.1 & 69.4 & 71.2 \\
        Our model ($\lambda = 0.5$) & \textbf{81.0} & \textbf{74.1} & \textbf{64.9} & \textbf{72.1} & \textbf{74.4} & \textbf{64.3} & \textbf{69.7} & \textbf{71.5} \\
        \bottomrule
    \end{tabular}
    }
\end{table*}

\section{Further Investigation}
\label{sec:furtherInvestigation}

%%%%%%%%%%%%%%%%%%%%
% \subsection{Ablation Study}
To better understand the mechanism of the proposed ELP, we demonstrate the necessity of key components ($\ie$, the $\lambda$ and the dynamic threshold strategy in self-training) of our model by conducting additional experiments.% for further investigation.

%%%%%%%%%%
\subsection{Dynamic Threshold vs Pre-defined Threshold.}
We compare two schemes of selecting clean data used in the aggregation phrase of intra-domain propagation with quantitative results.
The experiments are conducted on the tasks of ``\textit{painting to clipart}'', ``\textit{painting to real}'', and ``\textit{clipart to sketch}'' in the DomainNet dataset.

As shown in Fig.~\ref{fig:MedianVSOthers}, our median scheme achieves the best performance in all tasks.
Note that, for the clean classes obtained over $90\%$ accuracy, the proposed dynamic threshold strategy obtains the most numbers.
When conducting the fixed threshold experiments, we obverve that only two or three clean samples are obtained from some classes, and for some classes even no clean data is obtained. 
Apparently, using a fixed threshold cannot adapt to the complexity of classification task, the variance of the classes statistics, and additional domain shift in DA. 
The results indicate that using the dynamic threshold strategy does lay a better foundation for feature discriminability in the intra-domain propagation.

%%%%%%%%%%
\subsection{Ablation Study of Two-Stage Learning.}
We omits the Stage Two (described in Sec.~\ref{sec:LearningScheme}) in the proposed ELP framework and compare the classification accuracy with baseline (MME~\cite{saito2019semi}) and our full model.

As shown in Table~\ref{tab:AblationStudyOfTwoStageScheme}, we observe worse results on the classification accuracy than the results of full model but better results than the ones of baseline.
This demonstrate the necessity of both stages: we need the Stage One to learn a good initial representation space and the Stage Two to finetune the representation for further improving discriminability on the target domain.

%%%%%%%%%%
\subsection{Value of $\lambda$ in the intra-domain propagation.}
We train the model using different values of $\lambda$ in Eqs.~(\ref{equ:AssignPseudoLabel}) to investigate the effectiveness of our method in the intra-domain propagation stage.

As shown in Table~\ref{tab:AblationStudyTheAlphaInMixUp}, the \textit{random} row denotes that $\lambda$ is randomly obtained from the beta distribution by setting the corresponding hyper-parameter to $0.4$~\cite{zhang2018mixup}.
When $\lambda = 0$, we use the generated pseudo labels directly without balanced mixup to calculate the entropy loss, $\ie$ ignoring the label information of another mini-batch.
It is observed that the strategy of using random $\lambda$ values generally works inferior than using the proposed balance mixup, but better than without stage two and zero $\lambda$.
Particularly, the model achieves the best performance when $\lambda = 0.5$, while the performance decreases with the decrease of $\lambda$ ($\lambda < 0.5$).
These results demonstrate our claim that the pair of randomly selected mini-batches images from the same labeled target domain should be treated in an equivalent way.

% %%%%%%%%%%
% \subsection{X}

% %%%%%%%%%%
% \subsection{Investigation on Training Efficiency}
% To visualize the improved training efficiency promoted by the proposed TDP-Net method, we show the error curves of several methods on two datasets. The curves are exhibited in Fig~\ref{fig:VisualizationOfErrorCurves}.
% As we can observe that: 
% (1) The learning curves of methods in the UDA or SSDA setting generally demonstrate large variations, which calls for effective learning strategies.
% (2) Our method helps the CNN model achieve better optimal points, compared to the baseline approach.

% \input{figures/fig_VisualizationOfErrorCurves}
\section{Conclusion}

In this paper, we propose the ELP to reinforce the discriminability of learned representations in the target domain for semi-supervised domain adaptation.
To this end, we carefully design one inter-domain propagation and one intra-domain propagation.
First, ELP refines semantic-aware information propagation between two domains via a novel cycle discrepancy loss.
Second, ELP facilitates the feature discriminability in the target domain via an effective self-training scheme.
Through extensive experiments across three widely-used domain adaptation datasets and two typical adaptation cases, we demonstrate that the proposed ELP consistently helps promote the classification accuracies of state-of-the-art methods, in both semi-supervised domain adaptation and unsupervised domain adaptation.

For future work, we are interested in applying ELP in other usage scenarios, such as semantic segmentation and object detection.
Besides, we plan to extend ELP to handle other important and practical adaptation tasks, such as multi-source domain adaptation.

\bibliographystyle{ieee_fullname}
\bibliography{MAIN}

\begin{thebibliography}{10}\itemsep=-1pt

\bibitem{ben2010theory}
Shai Ben-David, John Blitzer, Koby Crammer, Alex Kulesza, Fernando Pereira, and
  Jennifer~Wortman Vaughan.
\newblock A theory of learning from different domains.
\newblock {\em Machine learning}, 79(1-2):151--175, 2010.

\bibitem{caron2018deep}
Mathilde Caron, Piotr Bojanowski, Armand Joulin, and Matthijs Douze.
\newblock Deep clustering for unsupervised learning of visual features.
\newblock In {\em Proceedings of the European Conference on Computer Vision
  (ECCV)}, pages 132--149, 2018.

\bibitem{chen2019transferability}
Xinyang Chen, Sinan Wang, Mingsheng Long, and Jianmin Wang.
\newblock Transferability vs. discriminability: Batch spectral penalization for
  adversarial domain adaptation.
\newblock In {\em International Conference on Machine Learning (ICML)}, pages
  1081--1090, 2019.

\bibitem{csurka2017domain}
Gabriela Csurka.
\newblock {\em Domain adaptation in computer vision applications}, volume~2.
\newblock Springer, 2017.

\bibitem{cui2020GVBG}
Shuhao Cui, Shuhui Wang, Junbao Zhuo, Chi Su, Qingming Huang, and Qi Tian.
\newblock Gradually vanishing bridge for adversarial domain adaptation.
\newblock In {\em Proceedings of the IEEE/CVF Conference on Computer Vision and
  Pattern Recognition (CVPR)}, pages 12455--12464, 2020.

\bibitem{deng2009imagenet}
Jia Deng, Wei Dong, Richard Socher, Li-Jia Li, Kai Li, and Li Fei-Fei.
\newblock Imagenet: A large-scale hierarchical image database.
\newblock In {\em 2009 IEEE Conference on Computer Vision and Pattern
  Recognition (CVPR)}, pages 248--255, 2009.

\bibitem{deng2018image}
Weijian Deng, Liang Zheng, Qixiang Ye, Guoliang Kang, Yi Yang, and Jianbin
  Jiao.
\newblock Image-image domain adaptation with preserved self-similarity and
  domain-dissimilarity for person re-identification.
\newblock In {\em Proceedings of the IEEE conference on computer vision and
  pattern recognition (CVPR)}, pages 994--1003, 2018.

\bibitem{dwibedi2019TCC}
Debidatta Dwibedi, Yusuf Aytar, Jonathan Tompson, Pierre Sermanet, and Andrew
  Zisserman.
\newblock Temporal cycle-consistency learning.
\newblock In {\em Proceedings of the IEEE Conference on Computer Vision and
  Pattern Recognition (CVPR)}, pages 1801--1810, 2019.

\bibitem{french2018SE}
Geoffrey French, Michal Mackiewicz, and Mark Fisher.
\newblock Self-ensembling for visual domain adaptation.
\newblock In {\em The International Conference on Learning Representations
  (ICLR)}, 2018.

\bibitem{ganin2015unsupervised}
Yaroslav Ganin and Victor Lempitsky.
\newblock Unsupervised domain adaptation by backpropagation.
\newblock In {\em Proceedings of the 32nd International Conference on Machine
  Learning (ICML)}, pages 1180--1189. JMLR. org, 2015.

\bibitem{grandvalet2005semi}
Yves Grandvalet and Yoshua Bengio.
\newblock Semi-supervised learning by entropy minimization.
\newblock In {\em Advances in Neural Information Processing Systems (NIPS)},
  pages 529--536, 2005.

\bibitem{han2018coteaching}
Bo Han, Quanming Yao, Xingrui Yu, Gang Niu, Miao Xu, Weihua Hu, Ivor Tsang, and
  Masashi Sugiyama.
\newblock Co-teaching: Robust training of deep neural networks with extremely
  noisy labels.
\newblock In {\em Advances in Neural Information Processing Systems (NeurIPS)},
  pages 8527--8537, 2018.

\bibitem{he2016deep}
Kaiming He, Xiangyu Zhang, Shaoqing Ren, and Jian Sun.
\newblock Deep residual learning for image recognition.
\newblock In {\em Proceedings of the IEEE Conference on Computer Vision and
  Pattern Recognition (CVPR)}, pages 770--778, 2016.

\bibitem{hoffman2018cycada}
Judy Hoffman, Eric Tzeng, Taesung Park, Jun-Yan Zhu, Phillip Isola, Kate
  Saenko, Alexei Efros, and Trevor Darrell.
\newblock Cycada: Cycle-consistent adversarial domain adaptation.
\newblock In {\em Proceedings of the 35th International Conference on Machine
  Learning (ICML)}, pages 1989--1998, 2018.

\bibitem{jiang2018mentornet}
Lu Jiang, Zhengyuan Zhou, Thomas Leung, Li-Jia Li, and Li Fei-Fei.
\newblock Mentornet: Learning data-driven curriculum for very deep neural
  networks on corrupted labels.
\newblock In {\em International Conference on Machine Learning (ICML)}, pages
  2304--2313, 2018.

\bibitem{kumar2010EA++}
Abhishek Kumar, Avishek Saha, and Hal Daume.
\newblock Co-regularization based semi-supervised domain adaptation.
\newblock In {\em Advances in Neural Information Processing Systems (NIPS)},
  pages 478--486, 2010.

\bibitem{kuznetsova2018openimage}
Alina Kuznetsova, Hassan Rom, Neil Alldrin, Jasper Uijlings, Ivan Krasin, Jordi
  Pont-Tuset, Shahab Kamali, Stefan Popov, Matteo Malloci, Tom Duerig, and
  Vittorio Ferrari.
\newblock The open images dataset v4: Unified image classification, object
  detection, and visual relationship detection at scale.
\newblock {\em arXiv preprint arXiv:1811.00982}, 2018.

\bibitem{lee2013pseudo}
Dong-Hyun Lee.
\newblock Pseudo-label: The simple and efficient semi-supervised learning
  method for deep neural networks.
\newblock In {\em Workshop on Challenges in Representation Learning, ICML},
  volume~3, page~2, 2013.

\bibitem{lee2019drop}
Seungmin Lee, Dongwan Kim, Namil Kim, and Seong-Gyun Jeong.
\newblock Drop to adapt: Learning discriminative features for unsupervised
  domain adaptation.
\newblock In {\em Proceedings of the IEEE International Conference on Computer
  Vision (ICCV)}, pages 91--100, 2019.

\bibitem{li2020online}
Da Li and Timothy Hospedales.
\newblock Online meta-learning for multi-source and semi-supervised domain
  adaptation.
\newblock In {\em Proceedings of the European conference on computer vision
  (ECCV)}, 2020.

\bibitem{li2018domain}
Shuang Li, Shiji Song, Gao Huang, Zhengming Ding, and Cheng Wu.
\newblock Domain invariant and class discriminative feature learning for visual
  domain adaptation.
\newblock In {\em IEEE Transactions on Image Processing}, pages 4260--4273.
  IEEE, 2018.

\bibitem{liang2020shot}
Jian Liang, Dapeng Hu, and Jiashi Feng.
\newblock Do we really need to access the source data? source hypothesis
  transfer for unsupervised domain adaptation.
\newblock In {\em International Conference on Machine Learning (ICML)}, July
  2020.

\bibitem{long2015mmd}
Mingsheng Long, Yue Cao, Jianmin Wang, and Michael Jordan.
\newblock Learning transferable features with deep adaptation networks.
\newblock In {\em International Conference on Machine Learning (ICML)}, pages
  97--105, 2015.

\bibitem{long2015learning}
Mingsheng Long, Yue Cao, Jianmin Wang, and Michael~I Jordan.
\newblock Learning transferable features with deep adaptation networks.
\newblock In {\em Proceedings of the 32nd International Conference on Machine
  Learning (ICML)}, pages 97--105. JMLR. org, 2015.

\bibitem{long2018conditional}
Mingsheng Long, Zhangjie Cao, Jianmin Wang, and Michael~I Jordan.
\newblock Conditional adversarial domain adaptation.
\newblock In {\em Advances in Neural Information Processing Systems (NeurIPS)},
  pages 1640--1650, 2018.

\bibitem{Lowe2004}
David~G. Lowe.
\newblock Distinctive image features from scale-invariant keypoints.
\newblock {\em Int. J. Comput. Vision}, 60:91--110, 2004.

\bibitem{maaten2008tsne}
Laurens van~der Maaten and Geoffrey Hinton.
\newblock Visualizing data using t-sne.
\newblock {\em Journal of machine learning research}, 9(Nov):2579--2605, 2008.

\bibitem{pan2019transferrableproto}
Yingwei Pan, Ting Yao, Yehao Li, Yu Wang, Chong-Wah Ngo, and Tao Mei.
\newblock Transferrable prototypical networks for unsupervised domain
  adaptation.
\newblock In {\em Proceedings of the IEEE Conference on Computer Vision and
  Pattern Recognition (CVPR)}, pages 2239--2247, 2019.

\bibitem{paszke2019pytorch}
Adam Paszke, Sam Gross, Francisco Massa, Adam Lerer, James Bradbury, Gregory
  Chanan, Trevor Killeen, Zeming Lin, Natalia Gimelshein, Luca Antiga, et~al.
\newblock Pytorch: An imperative style, high-performance deep learning library.
\newblock In {\em Advances in Neural Information Processing Systems (NeurIPS)},
  pages 8024--8035, 2019.

\bibitem{peng2019moment}
Xingchao Peng, Qinxun Bai, Xide Xia, Zijun Huang, Kate Saenko, and Bo Wang.
\newblock Moment matching for multi-source domain adaptation.
\newblock In {\em Proceedings of the IEEE International Conference on Computer
  Vision (ICCV)}, pages 1406--1415, 2019.

\bibitem{peng2017visda}
Xingchao Peng, Ben Usman, Neela Kaushik, Judy Hoffman, Dequan Wang, and Kate
  Saenko.
\newblock Visda: The visual domain adaptation challenge.
\newblock {\em arXiv preprint arXiv:1710.06924}, 2017.

\bibitem{ranjan2017l2}
Rajeev Ranjan, Carlos~D Castillo, and Rama Chellappa.
\newblock L2-constrained softmax loss for discriminative face verification.
\newblock {\em arXiv preprint arXiv:1703.09507}, 2017.

\bibitem{saito2019semi}
Kuniaki Saito, Donghyun Kim, Stan Sclaroff, Trevor Darrell, and Kate Saenko.
\newblock Semi-supervised domain adaptation via minimax entropy.
\newblock In {\em Proceedings of the IEEE International Conference on Computer
  Vision (ICCV)}, 2019.

\bibitem{saito2017asymmetric}
Kuniaki Saito, Yoshitaka Ushiku, and Tatsuya Harada.
\newblock Asymmetric tri-training for unsupervised domain adaptation.
\newblock In {\em Proceedings of the 34th International Conference on Machine
  Learning (ICML)}, pages 2988--2997. JMLR. org, 2017.

\bibitem{saito2017adversarial}
Kuniaki Saito, Yoshitaka Ushiku, Tatsuya Harada, and Kate Saenko.
\newblock Adversarial dropout regularization.
\newblock In {\em The International Conference on Learning Representations
  (ICLR)}, 2018.

\bibitem{saito2018MCD}
Kuniaki Saito, Kohei Watanabe, Yoshitaka Ushiku, and Tatsuya Harada.
\newblock Maximum classifier discrepancy for unsupervised domain adaptation.
\newblock In {\em Proceedings of the IEEE Conference on Computer Vision and
  Pattern Recognition (CVPR)}, pages 3723--3732, 2018.

\bibitem{saito2018maximum}
Kuniaki Saito, Kohei Watanabe, Yoshitaka Ushiku, and Tatsuya Harada.
\newblock Maximum classifier discrepancy for unsupervised domain adaptation.
\newblock In {\em Proceedings of the IEEE Conference on Computer Vision and
  Pattern Recognition (CVPR)}, pages 3723--3732, 2018.

\bibitem{Scharstein2002}
Daniel Scharstein and Richard Szeliski.
\newblock A taxonomy and evaluation of dense two-frame stereo correspondence
  algorithms.
\newblock {\em International Journal of Computer Vision}, 47(1-3):7--42, 2002.

\bibitem{simonyan2014very}
Karen Simonyan and Andrew Zisserman.
\newblock Very deep convolutional networks for large-scale image recognition.
\newblock {\em arXiv preprint arXiv:1409.1556}, 2014.

\bibitem{triguero2015self-labeled}
Isaac Triguero, Salvador Garc{\'\i}a, and Francisco Herrera.
\newblock Self-labeled techniques for semi-supervised learning: taxonomy,
  software and empirical study.
\newblock {\em Knowledge and Information systems}, 42(2):245--284, 2015.

\bibitem{venkateswara2017deep}
Hemanth Venkateswara, Jose Eusebio, Shayok Chakraborty, and Sethuraman
  Panchanathan.
\newblock Deep hashing network for unsupervised domain adaptation.
\newblock In {\em Proceedings of the IEEE Conference on Computer Vision and
  Pattern Recognition (CVPR)}, pages 5018--5027, 2017.

\bibitem{wu2018unsupervised}
Zhirong Wu, Yuanjun Xiong, Stella~X Yu, and Dahua Lin.
\newblock Unsupervised feature learning via non-parametric instance
  discrimination.
\newblock In {\em Proceedings of the IEEE Conference on Computer Vision and
  Pattern Recognition (CVPR)}, pages 3733--3742, 2018.

\bibitem{xu2020explore}
Chang-Dong Xu, Xing-Ran Zhao, Xin Jin, and Xiu-Shen Wei.
\newblock Exploring categorical regularization for domain adaptive object
  detection.
\newblock In {\em IEEE/CVF Conference on Computer Vision and Pattern
  Recognition (CVPR)}, June 2020.

\bibitem{xu2019adversarial}
Minghao Xu, Jian Zhang, Bingbing Ni, Teng Li, Chengjie Wang, Qi Tian, and
  Wenjun Zhang.
\newblock Adversarial domain adaptation with domain mixup.
\newblock In {\em The Thirty-Fourth AAAI Conference on Artificial
  Intelligence}, pages 6502--6509, 2020.

\bibitem{xu2019AFN}
Ruijia Xu, Guanbin Li, Jihan Yang, and Liang Lin.
\newblock Larger norm more transferable: An adaptive feature norm approach for
  unsupervised domain adaptation.
\newblock In {\em Proceedings of the IEEE International Conference on Computer
  Vision (ICCV)}, pages 1426--1435, 2019.

\bibitem{xu2019dSNE}
Xiang Xu, Xiong Zhou, Ragav Venkatesan, Gurumurthy Swaminathan, and Orchid
  Majumder.
\newblock d-sne: Domain adaptation using stochastic neighborhood embedding.
\newblock In {\em Proceedings of the IEEE Conference on Computer Vision and
  Pattern Recognition (CVPR)}, pages 2497--2506, 2019.

\bibitem{yao2015SDASL}
Ting Yao, Yingwei Pan, Chong-Wah Ngo, Houqiang Li, and Tao Mei.
\newblock Semi-supervised domain adaptation with subspace learning for visual
  recognition.
\newblock In {\em Proceedings of the IEEE conference on Computer Vision and
  Pattern Recognition (CVPR)}, pages 2142--2150, 2015.

\bibitem{zhan2020online}
Xiaohang Zhan, Jiahao Xie, Ziwei Liu, Yew-Soon Ong, and Chen~Change Loy.
\newblock Online deep clustering for unsupervised representation learning.
\newblock In {\em Proceedings of the IEEE/CVF Conference on Computer Vision and
  Pattern Recognition (CVPR)}, pages 6688--6697, 2020.

\bibitem{zhang2018mixup}
Hongyi Zhang, Moustapha Cisse, Yann~N Dauphin, and David Lopez-Paz.
\newblock mixup: Beyond empirical risk minimization.
\newblock {\em The International Conference on Learning Representations
  (ICLR)}, 2018.

\bibitem{zhou2020affinity}
Wei Zhou, Yukang Wang, Jiajia Chu, Jiehua Yang, Xiang Bai, and Yongchao Xu.
\newblock Affinity space adaptation for semantic segmentation across domains.
\newblock In {\em IEEE Transactions on Image Processing}. IEEE, 2020.

\bibitem{zhu2017unpaired}
Jun-Yan Zhu, Taesung Park, Phillip Isola, and Alexei~A Efros.
\newblock Unpaired image-to-image translation using cycle-consistent
  adversarial networks.
\newblock In {\em Proceedings of the IEEE International Conference on Computer
  Vision (ICCV)}, pages 2223--2232, 2017.

\bibitem{zou2019confidence}
Yang Zou, Zhiding Yu, Xiaofeng Liu, BVK Kumar, and Jinsong Wang.
\newblock Confidence regularized self-training.
\newblock In {\em Proceedings of the IEEE International Conference on Computer
  Vision (ICCV)}, pages 5982--5991, 2019.

\end{thebibliography}

\begin{IEEEbiography}
[{\includegraphics[width=1in, height=1.25in,clip,keepaspectratio]{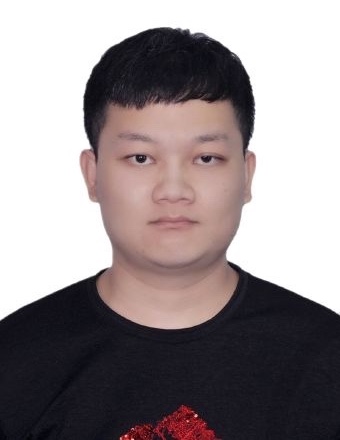}}]{Zhiyong Huang}
is a master student at the School of Control and Com-
puter Engineering, North China Electric Power University. He received his B.Eng. degree from North China Electric Power University in 2018. His research interest include single image super-resolution and domain adaptation.
\end{IEEEbiography}

\begin{IEEEbiography}
[{\includegraphics[width=1in, height=1.25in,clip,keepaspectratio]{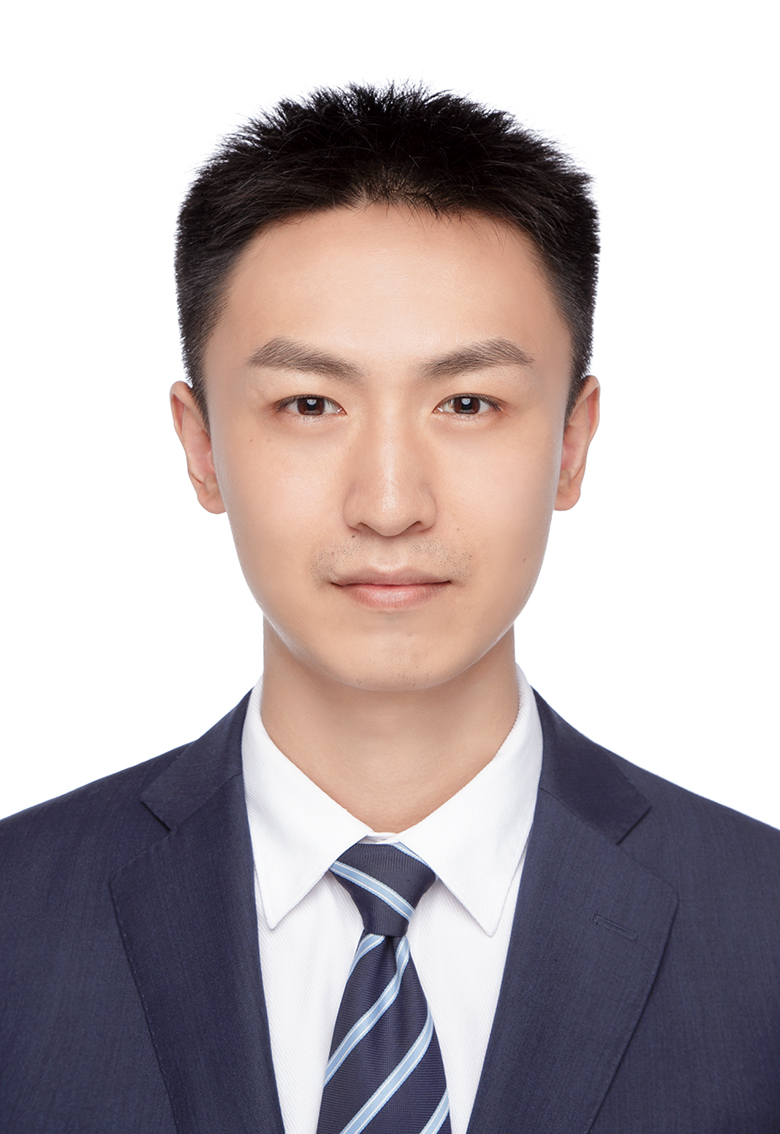}}]{Kekai Sheng} received his PhD. degree from National Laboratory of Pattern Recognition (NLPR), Institute of Automation, Chinese Academy of Sciences in 2019. He received his B.Eng. degree in Telecommunication Engineering from University of Science and Technology Beijing in 2014. He is
currently a researcher engineer at Youtu Lab, Tencent Inc. His research interests include image quality evaluation, domain adaptation, and AutoML.
\end{IEEEbiography}

\begin{IEEEbiography}
[{\includegraphics[width=1in, height=1.25in,clip,keepaspectratio]{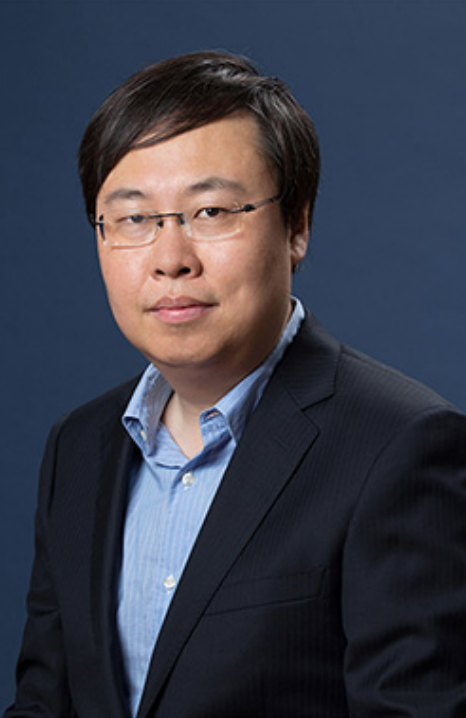}}]{Weiming Dong} is a Professor in the Sino-European Lab in Computer Science, Automation and Applied Mathematics (LIAMA) and National Laboratory of Pattern Recognition (NLPR) at Institute of Automation, Chinese Academy of Sciences. He received his BSc and MSc degrees in Computer Science in 2001 and 2004, both from Tsinghua University, China. He received his PhD in Computer Science from the University of Lorraine, France, in 2007. His research interests include visual media synthesis and image recognition. Weiming Dong is a member of the ACM and IEEE.
\end{IEEEbiography}

\begin{IEEEbiography}
[{\includegraphics[width=1in, height=1.25in,clip,keepaspectratio]{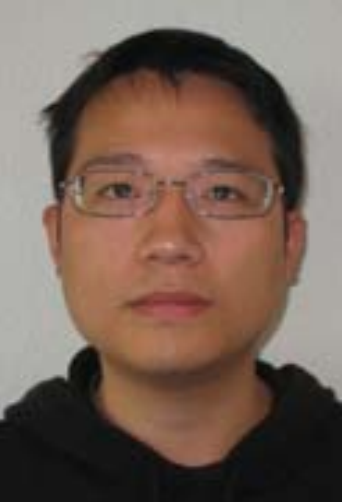}}]{Xing Mei} received his PhD degree from Institute of Automation, Chinese Academy of Sciences in 2009.
He is currently a software engineer at Bytedance Inc. His research interests include image processing, computer vision and computer graphics.
\end{IEEEbiography}

\begin{IEEEbiography}
[{\includegraphics[width=1in, height=1.25in,clip,keepaspectratio]{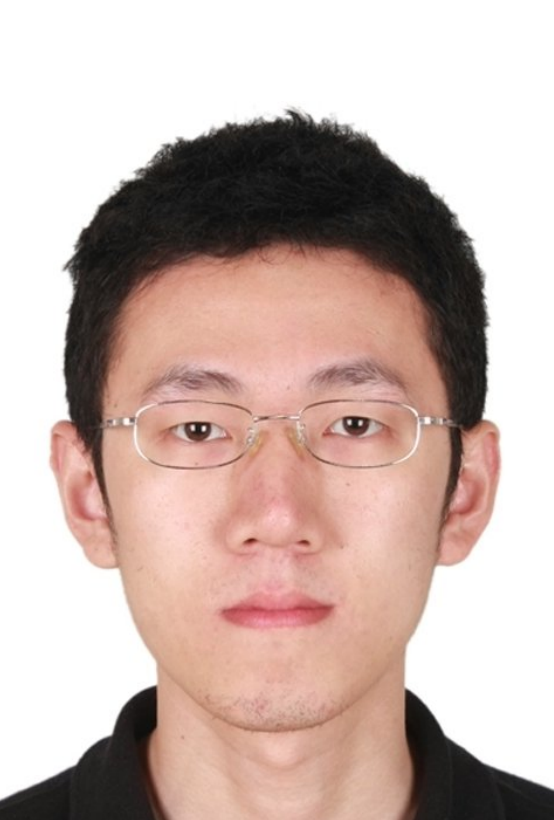}}]{Chongyang Ma} received B.S. degree from the Fundamental Science Class (Mathematics and Physics) of Tsinghua University in 2007 and PhD degree in Computer Science from the Institute for Advanced Study of Tsinghua University in 2012. He is currently a Research Lead at Kwai Inc. His research interests include computer graphics and computer vision.
\end{IEEEbiography}

% \begin{IEEEbiography}
% [{\includegraphics[width=1in, height=1.25in,clip,keepaspectratio]{images/authors/HaoleiYuan2019}}]{Haolei Yuan} received his M.Sc. degree in Signal Processing from Nanyang Technological University, Singapore, in 2012. He is currently a Researcher at Youtu Lab, Tencent. His research interests include image understanding and multi-modal learning.
% \end{IEEEbiography}

% \begin{IEEEbiography}
% [{\includegraphics[width=1in, height=1.25in,clip,keepaspectratio]{images/authors/XiaoweiGuo2020}}]{Xiaowei Guo} received his M.Sc. degrees in Mathematics from the School of Mathematics and Computational Science, Sun Yat-sen University, China, in 2005. He is currently a Research Lead at Youtu Lab, Tencent. His research interests include computer vision, deep learning, machine learning, etc.
% \end{IEEEbiography}

\begin{IEEEbiography}
[{\includegraphics[width=1in, height=1.25in,clip,keepaspectratio]{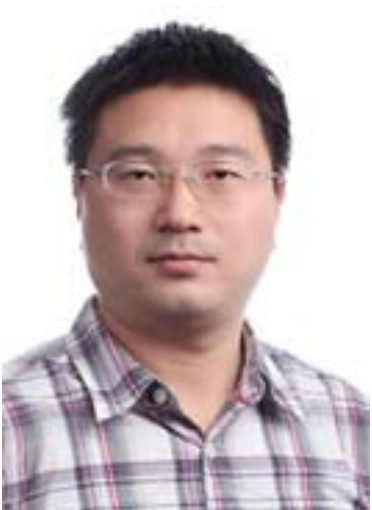}}]{Feiyue Huang} is the director of Youtu Lab, Tencent. He received his
BSc and PhD degrees in Computer Science in
2001 and 2008, both from Tsinghua University,
China. His research interests include image
understanding and face recognition.
\end{IEEEbiography}

\begin{IEEEbiography}
[{\includegraphics[width=1in, height=1.25in,clip,keepaspectratio]{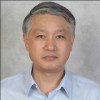}}]{Dengwen Zhou} is a Professor in the School of Control and Computer Engineering, North China Electric Power University, Beijing, China. He has long been engaged in research on image processing, including image denoising, image demosaicking, image interpolation and image super-resolution etc. Current research focuses on the applications based on neural networks and deep learning in image processing and computer vision.
\end{IEEEbiography}

\begin{IEEEbiography}
[{\includegraphics[width=1in, height=1.25in,clip,keepaspectratio]{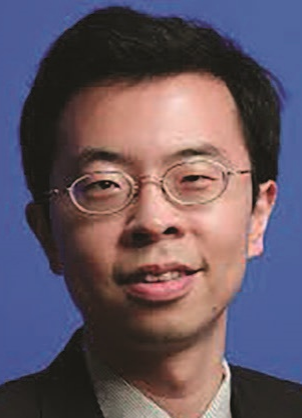}}]{Changsheng Xu} (Fellow, IEEE) is currently a Professor with the National Laboratory of Pattern Recognition, Institute of Automation, Chinese Academy of Sciences, and the Executive Director of the China-Singapore Institute of Digital Media. His research interests include multimedia content analysis/indexing/retrieval, pattern recognition, and computer vision. He has hold 30 granted/pending patents and published over 200 refereed research papers in these areas. He is a fellow of IAPR and an ACM Distinguished Scientist. He received the Best Associate Editor Award of \textit{ACM Transactions on Multimedia Computing, Communications and Applications} in 2012 and the Best Editorial Member Award of \textit{ACM/Springer Multimedia Systems Journal} in 2008. He served as a Program Chair of ACM Multimedia 2009. He has served as an Associate Editor, a Guest Editor, a General Chair, a Program Chair, an Area/Track Chair, a Special Session Organizer, a Session Chair, and a TPC member for over 20 IEEE and ACM prestigious multimedia journals, conferences, and workshops. He is an Associate Editor of \textit{ACM Transactions on Multimedia Computing, Communications and Applications} and \textit{ACM/Springer Multimedia Systems Journal}.
\end{IEEEbiography}

\end{document}